\documentclass[10pt,twocolumn,letterpaper]{article}

\usepackage{cvpr}              

\definecolor{cvprblue}{rgb}{0.21,0.49,0.74}
\usepackage[pagebackref,breaklinks,colorlinks,allcolors=cvprblue]{hyperref}

\usepackage{graphicx}
\usepackage{amsmath}
\usepackage{amssymb}
\usepackage{booktabs,caption}
\usepackage{multirow}
\usepackage{bm}
\usepackage{dsfont}
\usepackage[linesnumbered,ruled,vlined]{algorithm2e}
\usepackage{makecell}
\usepackage[flushleft]{threeparttable}
\usepackage{enumitem}
\usepackage[accsupp]{axessibility} 
\usepackage{algorithmic}
\usepackage[dvipsnames]{xcolor}
\definecolor{MyBlack}{HTML}{323A45}
\usepackage{placeins}
\usepackage{float}

\graphicspath{{figures/}}
\usepackage[pagebackref,breaklinks,colorlinks]{hyperref}

\def\eg{\emph{e.g.\ }}

\usepackage[capitalize]{cleveref}
\crefname{section}{Sec.}{Secs.}
\Crefname{section}{Section}{Sections}
\Crefname{table}{Table}{Tables}
\crefname{table}{Tab.}{Tabs.}

\def\paperID{170} 
\def\confName{CVPR}
\def\confYear{2026}

\title{The Third Challenge on Image Denoising at NTIRE 2026: Methods and Results}

\author{
Lei Sun$^*$ \and
Hang Guo$^*$ \and
Bin Ren$^*$ \and
Shaolin Su$^*$ \and
Xian Wang$^*$ \and
Danda Pani Paudel$^*$ \and
Luc Van Gool$^*$ \and
Radu Timofte$^*$ \and
Yawei Li$^*$ \and
Shiyan Jiang \and
Yukun Ma \and
Yansong Wang \and
Kairui Feng \and
Jingyuan Xie \and
Qi Zhu \and
Chunwei Tian \and
Jingyu Ma \and
Huiyuan Fu \and
Huadong Ma \and
Xiuhao Qiu \and
Xinchen Liu \and
Zhijie Ma \and
Jiawei Shi \and
Boqi Zhang \and
Dehao Feng \and
Yixiang Qiang \and
Zhe Yang \and
Hao Kang \and
Kun Liu \and
Arun Barkhanda \and
Wei Zhou \and
Hongyu Huang \and
Weijun Yuan \and
Xining Ge \and
Zhan Li \and
Gengjia Chang \and
Shuling Zheng \and
Feng Zhang \and
Zhiheng Fu \and
Naitik Pal \and
Ujjwal Mishra \and
Kailash A. Hambarde \and
Hugo Proença \and
Sachin Chaudhary \and
Praful Hambarde \and
Amit Shukla \and
Yash Arora \and
Aditya Arora \and
Yuedong Tan \and
Yuqi Li \and
Huiran Duan \and
Amruta Herur \and
Varsha I Pattanshetty \and
Sujatha C \and
Nikhil Akalwadi \and
Ramesh Ashok Tabib \and
Uma Mudenagudi \and
Shijun Shi \and
Jiangning Zhang \and
Yong Liu \and
Kai Hu \and
Jing Xu \and
Xianfang Zeng \and
Varsha C \and
A M Nandhana \and
Abhijith Sreeram \and
Aneesh Prian \and
Abhivanth Sivaprakash \and
Akash \and
Jiji C V \and
Param Pandya \and
Aagam Jain \and
Ved Patel \and
Milan Kumar Singh \and
Krish Parmar \and
Kishor Upla \and
Kiran Raja \and
Jiachen Tu \and
Guoyi Xu \and
Yaoxin Jiang \and
Jiajia Liu \and
Yaokun Shi \and
Tianyu Xiao \and
Changjia Wang \and
Yurao Deng \and
Tianjiao Wan \and
Cansu Korkmaz \and
Nancy Mehta \and
Shiqing Wang \and
Shaobo Xu \and
Yuan Xu \and
Kangwei Zhao \and
Yuqian Zhang \and
Yang Lu \and
Chaoyu Feng \and
Dongqing Zou \and
Lei Lei \and
Bilel Benjdira \and
Anas M. Ali \and
Wadii Boulila \and
Wenbin Wang \and
Xiaotong Luo \and
Yuan Gao \and
Wenjun Zeng
}

\begin{document}

\maketitle

\let\thefootnote\relax\footnotetext{$^*$ L. Sun (lei.sun@insait.ai, INSAIT, Sofia University, ``St. Kliment Ohridski''), H. Guo, B. Ren, S. Su, X. Wang, D. P. Paudel, L. Van Gool, R. Timofte, and Y. Li were the challenge organizers, while the other authors participated in the challenge. \\
Appendix~\ref{sec:teams} contains the authors' teams and affiliations.\\
NTIRE 2026 webpage: \url{https://cvlai.net/ntire/2026/}. \\
Code: \url{https://github.com/csguoh/NTIRE2026_Dn50_challenge}.
}

\begin{abstract}
This paper reports on the NTIRE 2026 Challenge on Image Denoising, specifically focusing on the high-noise regime ($\sigma = 50$). The competition investigates advanced neural architectures designed to restore high-fidelity details from images corrupted by additive white Gaussian noise (AWGN). Unlike constrained benchmarks, this track emphasizes peak quantitative performance, measured by Peak Signal-to-Noise Ratio (PSNR), without limitations on parameter count or computational overhead. By synthesizing contributions from 20 finalist teams out of 116 registrants, this report benchmarks the latest technical innovations and provides a comprehensive snapshot of the current state-of-the-art in unconstrained image restoration.
\end{abstract}

\section{Introduction}
\label{sec:introduction}
As a cornerstone of restorative vision, image denoising centers on the inverse mapping of corrupted observations back to their latent, noise-free state. This restoration is inherently ill-posed because real-world acquisition pipelines introduce a complex mixture of stochastic processes, encompassing additive Gaussian and signal-dependent Poisson noise as well as structured quantization errors such as JPEG artifacts. The demand for robust denoising algorithms remains critical across diverse domains, particularly in high-end computational photography, diagnostic medical imaging, and satellite-based remote sensing. Consequently, there is a persistent need for models that offer both superior generalization and high-fidelity reconstruction~\cite{gu2019brief,zhang2017beyond,qian2025learning}.

To catalyze innovation within the image restoration community, this competition is dedicated to fostering the next generation of denoising frameworks. We adopt the Additive White Gaussian Noise (AWGN) model as our primary evaluative benchmark, as it provides a robust and standardized environment for the fair quantitative assessment of diverse architectural designs. By utilizing this classical degradation model, the challenge ensures that performance gains can be directly attributed to methodological advancements.

Held in conjunction with the New Trends in Image Restoration and Enhancement (NTIRE) 2026 workshop, this competition follows a series of successful challenges that have significantly advanced the state-of-the-art in computational photography. As a premier venue for benchmarking image restoration algorithms, the NTIRE workshop continues to push the boundaries of the field~\cite{sun2025tenth,ntire2025event,ntire2025face,ntire2025srx4,ntire2025esr}. This year's iteration specifically addresses the persistent challenge of image denoising under a high-noise regime ($\sigma=50$), aiming to evaluate the robustness of modern architectures in extreme restoration scenarios. By benchmarking models against a standardized AWGN degradation, the challenge aims to catalyze the development of next-generation restoration topologies. Beyond identifying top-performing architectures, this initiative serves as a platform for analyzing emerging technical shifts, ultimately providing a comprehensive roadmap for future research in the field.

This challenge is one of the challenges associated with the NTIRE 2026 Workshop~\footnote{\url{https://www.cvlai.net/ntire/2026/}} on:
deepfake detection~\cite{ntire26deepfake}, 
high-resolution depth~\cite{ntire26hrdepth},
multi-exposure image fusion~\cite{ntire26raim_fusion}, 
AI flash portrait~\cite{ntire26raim_portrait}, 
professional image quality assessment~\cite{ntire26raim_piqa},
light field super-resolution~\cite{ntire26lightsr},
3D content super-resolution~\cite{ntire263dsr},
bitstream-corrupted video restoration~\cite{ntire26videores},
X-AIGC quality assessment~\cite{ntire26XAIGCqa},
shadow removal~\cite{ntire26shadow},
ambient lighting normalization~\cite{ntire26lightnorm},
controllable Bokeh rendering~\cite{ntire26bokeh},
rip current detection and segmentation~\cite{ntire26ripdetseg},
low light image enhancement~\cite{ntire26llie},
high FPS video frame interpolation~\cite{ntire26highfps},
Night-time dehazing~\cite{ntire26nthaze,ntire26nthaze_rep},
learned ISP with unpaired data~\cite{ntire26isp},
short-form UGC video restoration~\cite{ntire26ugcvideo},
raindrop removal for dual-focused images~\cite{ntire26dual_focus},
image super-resolution (x4)~\cite{ntire26srx4},
photography retouching transfer~\cite{ntire26retouching},
mobile real-word super-resolution~\cite{ntire26rwsr},
remote sensing infrared super-resolution~\cite{ntire26rsirsr},
AI-Generated image detection~\cite{ntire26aigendet},
cross-domain few-shot object detection~\cite{ntire26cdfsod},
financial receipt restoration and reasoning~\cite{ntire26finrec},
real-world face restoration~\cite{ntire26faceres},
reflection removal~\cite{ntire26reflection},
anomaly detection of face enhancement~\cite{ntire26anomalydet},
video saliency prediction~\cite{ntire26videosal},
efficient super-resolution~\cite{ntire26effsr},
3d restoration and reconstruction in adverse conditions~\cite{ntire26realx3d},
image denoising~\cite{ntire26denoising},
blind computational aberration correction~\cite{ntire26aberration},
event-based image deblurring~\cite{ntire26eventblurr},
efficient burst HDR and restoration~\cite{ntire26bursthdr},
low-light enhancement: `twilight cowboy'~\cite{ntire26twilight},
and efficient low light image enhancement~\cite{ntire26effllie}.



\section{NTIRE 2026 Image Denoising Challenge}
The mission of this challenge is structured around three primary pillars: (i) to catalyze methodological breakthroughs in image restoration; (ii) to facilitate a rigorous and equitable benchmarking process for diverse denoising paradigms; and (iii) to bridge the gap between academia and industry by fostering a synergistic platform for knowledge exchange and technical collaboration.

Subsequent sections delineate the organizational framework of the challenge, covering the dataset characteristics, evaluation protocols, final rankings, and a synthesis of the architectural innovations introduced by the participating teams. By providing this unified benchmark, we aim to redefine the performance frontiers of modern denoising and stimulate sustainable innovation within the community.

\section{Datasets}
\label{sec:datasets}
The same as last year~\cite{ntire2025denoising}, the challenge utilizes two benchmark datasets for training and evaluation: \textbf{DIV2K}~\cite{agustsson2017ntire} and \textbf{LSDIR}~\cite{lilsdir}.

\noindent{\textbf{DIV2K Dataset:}} This dataset features 1,000 diverse RGB images at 2K resolution. The data distribution follows a standard split of 800 images for training, 100 for validation, and 100 for testing purposes.

\noindent{\textbf{LSDIR Dataset:}} Comprising 86,991 high-quality, high-resolution samples, this dataset is allocated as follows: 84,991 images for training, 1,000 for validation, and 1,000 for testing.

\noindent{\textbf{Data Usage Protocol:}} Participants were granted access to the combined training sets from both DIV2K and LSDIR. For the validation stage, the 100 images from the DIV2K validation partition were provided. The final evaluation during the test phase was performed on a total of 200 images, consisting of 100 samples from the DIV2K test set and an additional 100 from the LSDIR test set. To maintain a rigorous and unbiased assessment, the ground-truth noise-free images for the testing phase remained strictly sequestered and were not accessible to participants at any point during the competition.

\subsection{Competition Framework and Tracks}
The core objective is the engineering of neural architectures capable of producing high-fidelity denoising outputs, with performance primarily quantified via the Peak Signal-to-Noise Ratio (PSNR) metric.

\medskip
\noindent{\textbf{Challenge Phases}}
\begin{enumerate}
    \item \textit{Development and Validation}: Participants were furnished with 800 clean samples and 100 clean/noisy image pairs from the DIV2K corpus, supplemented by 84,991 clean images from the LSDIR dataset. During the training stage, noisy observations were synthesized by injecting Additive White Gaussian Noise (AWGN) at a consistent level of $\sigma = 50$. The CodaLab evaluation server served as a real-time benchmarking platform, allowing teams to submit their restored outputs and receive instantaneous PSNR feedback to guide their model optimization.
    \item \textit{Final Evaluation}: In the conclusive testing stage, participants were granted access to a test set comprising 100 noisy images from DIV2K and 100 from LSDIR. To ensure the integrity of the results, the corresponding ground-truth references were strictly withheld. Submissions required the restored images, the underlying source code, and a descriptive factsheet. The organizing committee subsequently performed a verification of the codebases to regenerate and validate the final results, which were disclosed to all participants upon the competition's completion.
\end{enumerate}

\medskip
\noindent{\textbf{Quantitative Assessment and Evaluation Metrics}}
The primary mandate of this challenge is to maximize the precision of image restoration under high-noise conditions. To this end, the final leaderboard is established based on the Peak Signal-to-Noise Ratio (PSNR) and Structural Similarity Index (SSIM), computed across the 200-image ensemble test set. For transparency and reproducibility, the evaluation protocol, along with the participants' source code and pre-trained weights, is hosted at \url{https://github.com/csguoh/NTIRE2026_Dn50_challenge}. Crucially, the ranking criteria remain strictly performance-driven and are agnostic to architectural complexity or computational footprint; therefore, model size and inference efficiency were not factored into the official results.

\section{Challenge Results}
\label{sec:results}

The final leaderboard and quantitative performance of the participating teams are summarized in Table~\ref{tab:rank}. For an in-depth technical analysis of each submission, readers are referred to the methodology summaries in Sec.~\ref{sec:methods_and_teams}, while the complete roster of participants and their respective affiliations are detailed in Appendix~\ref{sec:teams}. 

Notably, the competition was exceptionally fierce among the top contenders, with the leading four teams separated by less than $0.03$ dB. \textbf{BuptMM} secured the first place with a PSNR of 29.90 dB, followed closely by \textbf{I2WM\&JNU} (29.89 dB), \textbf{MIIE} (29.873 dB), and \textbf{Titans} (29.871 dB). These marginal differences underscore the intense quest for reconstruction fidelity and represent the current frontier in high-performance image denoising under severe noise conditions ($\sigma=50$).

\begin{table}[t]
    \centering
    \begin{tabular}{l|c|cc}
    \hline
    Team & Rank & PSNR (primary) & SSIM \\ \hline
    BuptMM & 1 & 29.90 & - \\
    I2WM\&JNU & 2 & 29.89 & - \\
    MIIE & 3 & 29.873 & - \\
    Titans & 4 & 29.871 & - \\
    YuFans & 5 & 29.80 & - \\
    Noice & 6 & 29.77 & - \\
    Perceptual Vision Team & 7 & 29.61 & - \\
    NextAI & 8 & 29.53 & - \\
    SNUC & 9 & 29.51 & - \\
    PSU & 10 & 29.49 & - \\
    NTR & 11 & 29.36 & - \\
    tarrya & 12 & 29.00 & - \\
    ML\_SVNIT & 13 & 28.99 & - \\
    KLETech-CEVI & 14 & 28.98 & - \\
    Variational Vision & 15 & 28.95 & - \\
    nudt\_bestin\_ntire & 16 & 28.84 & - \\
    APRIL-AIGC & 17 & 28.24 & - \\
    mandalinadagi & 18 & 27.29 & - \\
    CV\_SVNIT & 19 & 26.97 & - \\
    WUrbane & 20 & 26.93 & - \\ \hline
    \end{tabular}
    \caption{Results of NTIRE 2026 Image Denoising Challenge. PSNR and SSIM scores are measured on the 200 test images from DIV2K test set and LSDIR test set. Team rankings are based primarily on PSNR.}
    \label{tab:rank}
\end{table}

\subsection{Participants}
\label{sec:participants}
This year, the challenge attracted 116 registered participants, with 20 teams successfully submitting valid results. In the following sections, we provide a high-level summary of the participants' approaches, while a more detailed description of each method is presented in Section \ref{sec:methods_and_teams}.

\subsection{Main Ideas and Architectures}
\label{sec:main_ideas}
During the challenge, participants implemented a range of novel techniques to enhance image denoising performance. Below, we highlight some of the fundamental strategies adopted by the leading teams.

\begin{enumerate}
    \item \textbf{Exploitation of Mamba and Transformers.} Advanced architectures like {MambaIR} and {Restormer} remain the backbones of choice. {I2WM\&JNU} (Rank 2) utilized Restormer's multi-Dconv head transposed attention (MDTA) to capture global context with linear complexity. {Titans} (Rank 4) introduced {UnifyFormer}, which augments self-attention with multi-scale local aggregation through separable convolutions of varying kernel sizes.
    
    \item \textbf{Large-scale and High-quality Data Strategies.} Data diversity and quality were critical for performance gains. {I2WM\&JNU} constructed an extensive dataset of over 140k images from sources like DIV2K, LSDIR, Flickr2K, and LIU4K-v2. Both {BuptMM} and {MIIE} emphasized a multi-stage training process where the model is first pre-trained on large-scale data and then fine-tuned on filtered, high-quality subsets to refine texture and edge sharpness.
    
    \item \textbf{Progressive Learning and Sophisticated Loss Functions.} A common theme among the leaders was the use of progressive training, where patch sizes are gradually increased (\eg, from 128 to 768) to improve model robustness and receptive field. To move beyond standard pixel-wise losses, {MIIE} introduced frequency-domain constraints like ``Stationary Wavelet Transform (SWT) Loss'' to enhance high-frequency detail recovery.
    
    \item \textbf{Inference-time Enhancement (TTA).} Test-time strategies were essential for squeezing out the final decibels of PSNR. Most top teams adopted ``Self-ensemble'' (augmenting inputs via rotation/flipping)~\cite{timofte2016seven} and ``Test-time Local Converter (TLC)'' to refine local details~\cite{chu2021revisiting}. Additionally, overlapping tile inference with sophisticated blending (e.g., cosine window splicing) was used to handle high-resolution images without artifacts.
    
\end{enumerate}

\subsection{Ensuring Fairness and Competition Integrity}
\label{sec:fairness}

To maintain a rigorous and equitable environment for all participants, specific protocols were enforced regarding data utilization and model optimization. The regulatory framework for the challenge is outlined as follows:

\begin{enumerate}
    \item \textbf{Data Augmentation and External Corpora:} Participants were permitted to supplement the official training data with external repositories, such as Flickr2K. Furthermore, the application of sophisticated data augmentation strategies was encouraged and deemed consistent with the competitive standards.
    \item \textbf{Strict Data Partitioning:} To preserve the validity of the evaluation process, any form of training on the DIV2K validation partition, including both high-resolution and degraded counterparts, was strictly forbidden. This measure ensures that the validation phase remains a reliable indicator of a model's generalization capabilities.
    \item \textbf{Test Set Confidentiality:} Utilizing images from the DIV2K test set during the training or fine-tuning stages was prohibited. These constraints were established to prevent any potential data leakage or over-fitting to the specific evaluation distribution.
\end{enumerate}


\section{Challenge Methods and Teams}
\label{sec:methods_and_teams}

\subsection{NextAI}
Our overall network architecture adopts a structure similar to U-Net, consisting of an encoder, a middle module, and a decoder, where multi-head self-attention is utilized as the basic building block. Specifically, the encoder contains 4 downsampling layers, with the number of blocks in each layer being [2, 2, 3, 3]. The middle module is composed of 3 blocks. The decoder incorporates 4 upsampling layers, and the number of blocks per layer is set to [3, 3, 2, 2].

For the training phase, we conduct experiments on the DIV2K and LSDIR datasets. Data augmentation strategies including horizontal flipping, vertical flipping, and random rotation are applied. Subsequently, patches are cropped from the augmented images, and noise is added to these patches to construct training pairs. The training process is divided into three stages: the first stage trains the model for 50 epochs with a resolution of 128, the second stage trains for 30 epochs with a resolution of 256, and the third stage trains for 20 epochs with a resolution of 512. We employ the L1 loss and L2 loss as the loss functions. The optimizer and learning rate scheduler used during training were AdamW and CosineAnnealingLR, respectively.

In the inference phase, we employ a self-ensemble strategy~\cite{timofte2016seven} and selectively adopt the TLC method \cite{chu2021revisiting} to further boost model performance.
\subsection{Perceptual Vision Team}
\begin{figure*}[h]
    \centering
    \includegraphics[width=1.0\linewidth]{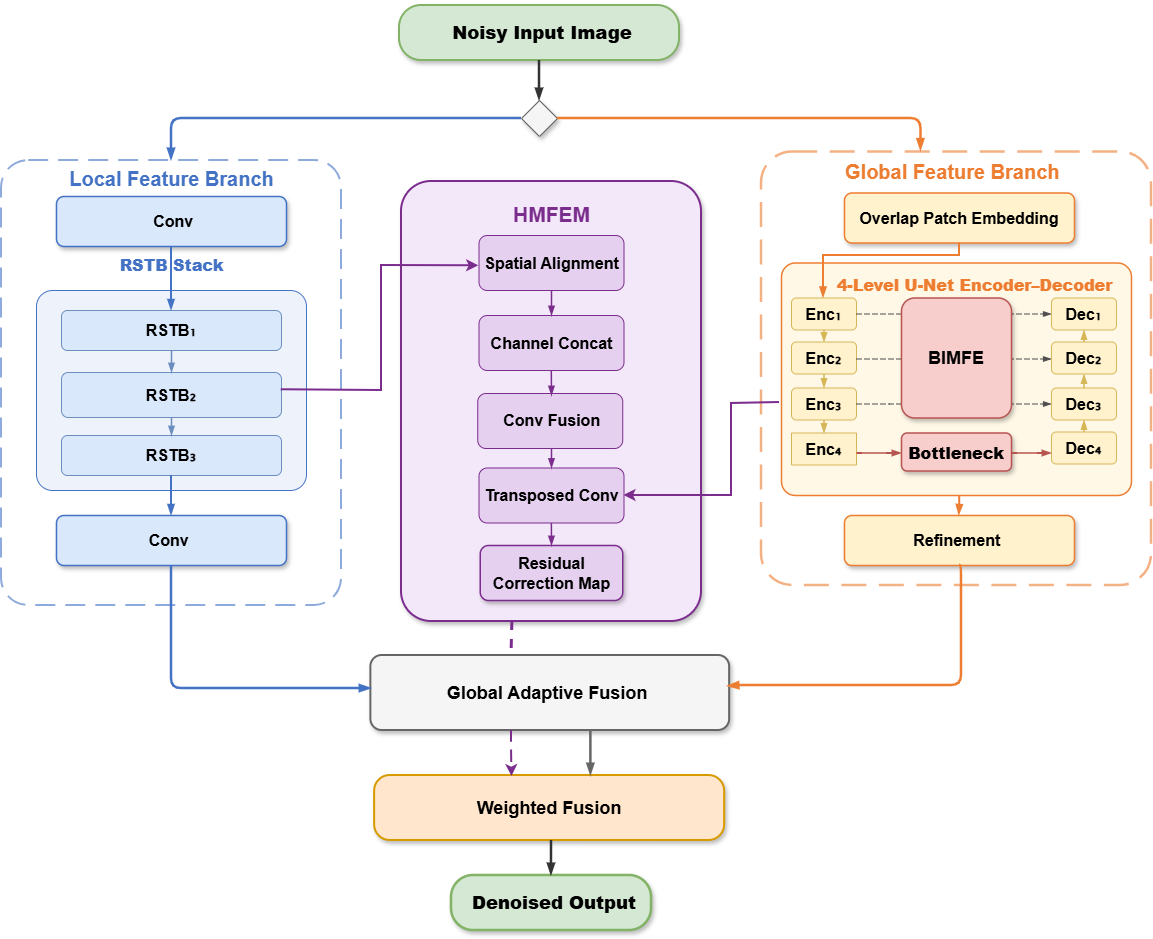}
    \caption{Overall architecture of the DualExNet proposed by Perceptual Vision Team.}
    \label{fig:arch}
\end{figure*}

\paragraph{}
As illustrated in \cref{fig:arch}, a new Dual-Branch Exchange Network (DualExNet)  is proposed. In consideration of the double dependency of the image denoising task on local high-frequency information and global low-frequency information, a parallel dual-branch architecture is employed in the network. 

The forward propagation of the proposed model employs a parallel dual-stream architecture: the local feature branch extracts fine-grained local features via Residual Swin Transformer Blocks (RSTB)\cite{liang2021swinir}, while the global feature branch captures global context utilizing a U-Net structure\cite{zamir2022restormer}. Notably, the Bi-directional Multi-scale Feature Exchange Module (BIMFE) module is integrated at the skip connections between the encoder and decoder, facilitating the adaptive fusion and residual enhancement of cross-scale (Top-Down/Bottom-Up) features through learnable gating mechanisms. Subsequently, the multi-scale intermediate features from both branches are fed into the Hybrid Multi-scale Feature Exchange Module (HMFEM) module, where they undergo spatial alignment, channel concatenation, and progressive upsampling reconstruction to generate a global residual correction map. The final output is obtained by dynamically weighting the predictions of the dual branches using pixel-level weights generated by an adaptive gating network, followed by the addition of the HMFEM residual. Furthermore, a self-ensemble strategy encompassing 8 geometric transformations~\cite{timofte2016seven} is introduced during the inference phase, which further boosts the denoising performance and robustness of the model through multi-view prediction fusion.

\subsubsection{Bi-directional Multi-scale Feature Exchange Module}
\paragraph{}
In traditional hierarchical feature reconstruction, single-scale feature fusion often leads to local loss of multi-scale contextual information. To introduce cross-scale information flow and enhance inter-layer feature interaction, we integrate BIMFE into the skip connections of the global feature branch.

For the BIMFE module at level $i$, its input is defined as the current level feature $X_{main}$, the high-resolution feature $X_{prev}$ from the upper-level encoder, and the low-resolution feature $X_{next}$ from the lower-level decoder. Its multi-scale fusion mechanism can be formally expressed as:

\begin{equation}
\begin{split}
F_{fusion} &= X_{main} + \alpha_{down} \cdot \mathcal{D}(X_{prev}) \\
           &+ \alpha_{up} \cdot \mathcal{U}(X_{next})
\end{split}
\end{equation}
where $\mathcal{D}(\cdot)$ denotes the downsampling operation based on $3 \times 3$ convolution (stride 2); $\mathcal{U}(\cdot)$ denotes the upsampling operation composed of $1 \times 1$ convolution and pixel shuffle. $\alpha_{down}$ and $\alpha_{up}$ are learnable gating weights initialized to 0.5, allowing the network to adaptively optimize the contribution of cross-scale feature flows during training.

To ensure the stability of gradient propagation in the network backbone (i.e., the integrity of the original feature flow) while incorporating multi-scale information, the fused feature $F_{fusion}$ needs to sequentially pass through Layer Normalization (LayerNorm) and Attention mechanisms for feature filtering. It is then superimposed on the current main branch feature $X_{main}$ in the form of a residual:
\begin{equation}
 X'_{out} = X_{main} + \text{Attention}(\text{LN}(F_{fusion})) 
\end{equation}
The final feature undergoes non-linear mapping through a Feed-Forward Network (FFN):
\begin{equation}
 X_{out} = X'_{out} + \text{FFN}(\text{LN}(X'_{out})) 
\end{equation}
where $X'_{out}$ represents the intermediate feature after attention modulation, and $X_{out}$ is the final output of the BIMFE block.

\subsubsection{Hybrid Multi-scale Feature Exchange Module}
\paragraph{}
The core purpose of HMFEM is to eliminate the semantic differences between the local feature and global feature branches at the feature level, achieving deep fusion of local and global features. This module receives the hierarchical feature set $\{S_0, S_1, S_2\}$ from local feature branch and the corresponding feature set $\{R_0, R_1, R_2\}$ from global feature branch.

Due to differences in spatial resolution and channel dimensions, HMFEM first utilizes $1 \times 1$ convolution to perform channel projection  on $S_i$, and performs spatial alignment via bilinear interpolation. Subsequently, concatenation is performed in the channel dimension, and deep aggregation is achieved through a $3 \times 3$ convolutional layer with GELU activation:
\begin{equation}
M_i = \text{GELU}\Big(\text{Conv}_{3\times3}\big([ \text{Proj}(S_i), R_i ]\big)\Big)
\end{equation}
where $M_i$ denotes the fused feature map at scale $i$, $\text{Proj}(\cdot)$ represents the channel projection and spatial alignment operation, and $[\cdot, \cdot]$ denotes the concatenation operation. The generated multi-scale feature representations $\{M_0, M_1, M_2\}$ will be progressively upsampled and fused through a bottom-up path based on transposed convolution . Finally, an output projection layer maps it to the residual correction map $R_{hmfem}$ in the global image space.

\subsubsection{Global Adaptive Fusion Module}
\paragraph{}
To obtain the optimal output reconstruction result, this model abandons the naive average pooling strategy and designs an adaptive gating fusion mechanism based on feature content, which is employed in both training and inference phases. Specifically, the dynamic gating map is first generated by concatenating the two branch prediction maps $P_{\text{swin}}$ and $P_{\text{restormer}}$ along the channel dimension. We then obtain the pixel-level dynamic gating weight map $G \in [0, 1]$ via a lightweight network with two convolutional layers and GELU, Sigmoid activations. Then, spatial weighted fusion is implemented by employing this weight map to conduct soft-gating combination on the two sets of predictions, formulated as:

\begin{equation}
P_{\text{gated}} = G \odot P_{\text{swin}} + (1 - G) \odot P_{\text{restormer}}
\end{equation}
where $\odot$ denotes element-wise multiplication. Finally, we perform high-frequency residual compensation by accumulating the residual features from the High-Frequency Multi-Feature Enhancement Module (HMFEM) to the fused image, yielding the final denoised output with refined texture.

\subsubsection{Train Strategy}
\paragraph{}
We train our model on a combined dataset comprising 800 high-resolution images from DIV2K \cite{agustsson2017ntire} and 84,991 images from LSDIR \cite{li2023lsdir}, with a total of 85,791 training pairs.This large-scale training data helps ensure robust generalization across diverse scenes. The training process is supervised using the Mean Absolute Error (MAE), i.e., $\mathcal{L}_1$ Loss, which effectively constrains pixel-level reconstruction fidelity. We employ the AdamW optimizer with default parameters $\beta_1 = 0.9$, $\beta_2 = 0.999$, and a weight decay of $1 \times 10^{-4}$ to minimize the loss function. The initial learning rate is set to $1 \times 10^{-4}$ and is scheduled using the Cosine Annealing strategy, gradually decaying to a minimum of $1 \times 10^{-7}$ over the entire training cycle.During the inference phase, we employ a self-ensemble strategy~\cite{timofte2016seven} to improve denoising performance. We perform 8 geometric transformations on the input image, consisting of horizontal/vertical flipping and rotation at $0^\circ, 90^\circ, 180^\circ, 270^\circ$. The final denoised result is obtained by averaging the inversely transformed predictions of all augmented images,.

\subsubsection{Experimental Results}
\paragraph{}
The objective quality evaluation of the model employs Peak Signal-to-Noise Ratio (PSNR) and Structural Similarity (SSIM) as core metrics. Based on the validation set and test set provided officially by the NTIRE 2026 Challenge, the quantitative performance of this model is shown in \cref{tab:results}:

\begin{table}[h]
\centering
\caption{Quantitative performance on NTIRE 2026 dataset.}
\label{tab:results}
\begin{tabular}{lcc}
\toprule
Dataset (NTIRE 2026) & PSNR (dB) & SSIM \\
\midrule
Validation Set & 28.10 & 0.79 \\
Test Set & 29.61 & 0.86 \\
\bottomrule
\end{tabular}
\end{table}

\subsection{BuptMM}
\paragraph{}
In recent years, the transformer architecture remains a leading method in denoising tasks. Despite the strong perceptual quality of generative models, the main goal of denoising is to restore the original image rather than guess possible details. Therefore, models based on the Diffusion architecture still lag behind leading architectures such as Restormer \cite{zamir2022restormer}, HAT \cite{hat}, and MambaIR \cite{MambaIR} in this field. Our method trained three models separately and observed their results. We found that HAT excels at restoring edge regions, MambaIR maintains better fidelity in smooth regions, while Restormer is more balanced. Therefore, after obtaining the three results separately, we fused them to achieve the entire denoising task and proposed the hybrid denoising Unit(HDU). However, in order to obtain higher-quality results, our model demands higher computational power and inference time, which may become unfriendly to some users.

\begin{figure*}[h]
    \centering
    \includegraphics[width=1.0\linewidth]{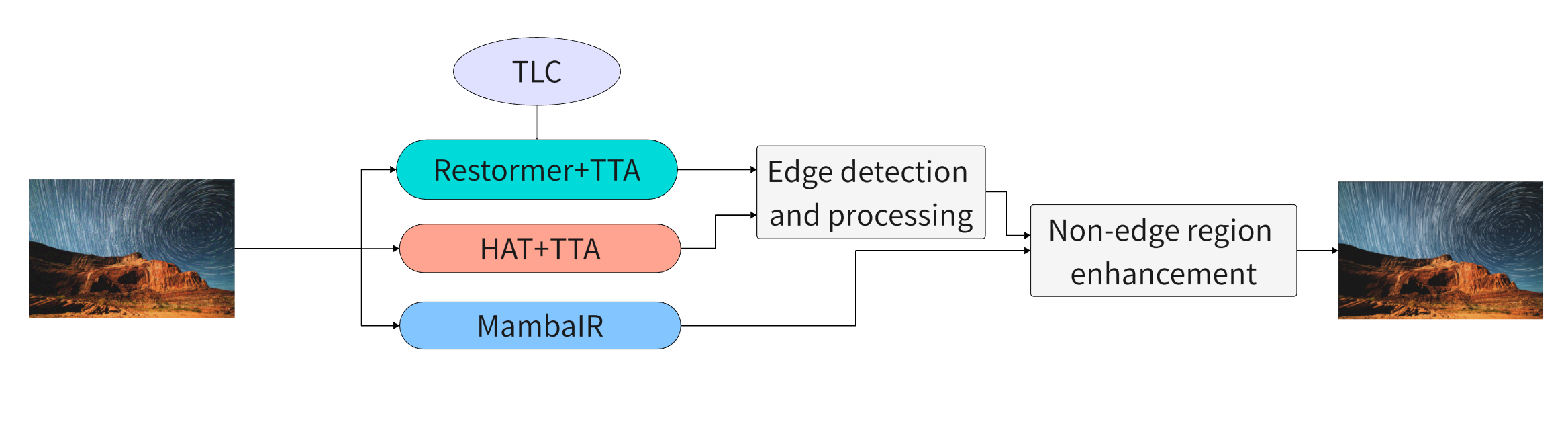}
    \caption{Overall architecture of HDU proposed by team BuptMM.}
    \label{fig:BuptMM}
\end{figure*}

As shown in \cref{fig:BuptMM}, our network can generally be divided into two stages. In the first stage, we use DIV2K \cite{agustsson2017ntire} and LSDIR \cite{li2023lsdir} and 25K high quality images from the Unsplash website to train Restormer, HAT and MambaIR respectively, and then enhance the ability of Restormer through TLC \cite{chu2021revisiting} technology and use TTA for both restormer and HAT during their reasoning stage. In the second stage, we first use the Canny operator to perform edge detection on the images processed by the two models. We take an OR operation on the two edge images and then XOR the result with the edge of HAT to obtain the edge difference between the two images. For this part of the edge difference, we use the result obtained by HAT as the standard for preservation. Finally, we take the average of the other edge pixels of HAT and Restormer to obtain the final edge result. For other non-edge pixels, we noticed that although MambaIR's overall PSNR and SSIM are not as good as the aforementioned two models, for non-edge positions, using a proportion of 0.1\%-1\% and the mean values of Restormer and HAT at these pixels can achieve a PSNR improvement of 0.05-0.15 on the validation set. We ultimately adopted a proportion of 0.1\% for fusion and obtained the final image.

We used the DIV2K and LSDIR and Unsplash datasets to train both the Restormer and HAT and MambaIR simultaneously on 8 H200 GPUs. We employed a progressive training strategy for the Restormer, where the image block size increased from 128 to 384 with a step size of 64. We also used a fixed size training strategy for the HAT, with a patch size of 512 and a fixed size 256 for MambaIR. We achieved a score of 30.70 on the DIV2K validation set.

\subsection{Variational\_Vision}
\subsubsection{General Method Description}
\begin{figure*}[h]
    \centering
    \includegraphics[width=\linewidth]{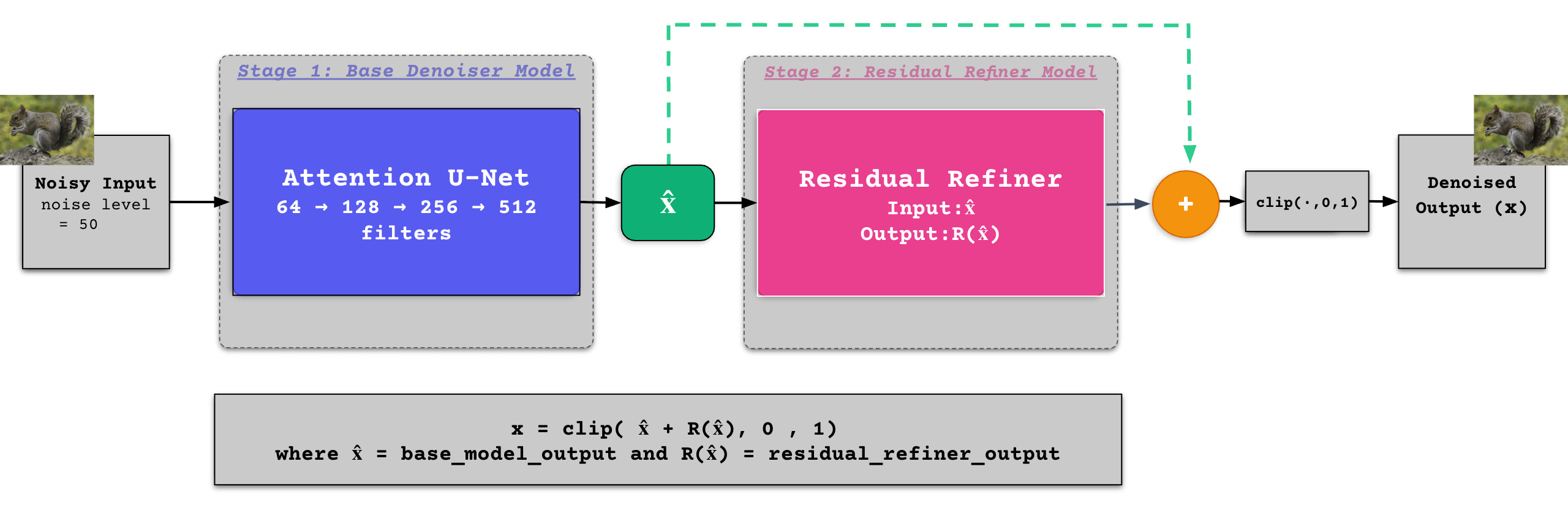}
    \caption{Overview of the two-stage denoising pipeline proposed by team Variational\_Vision. The base Attention U-Net produces an initial denoised image $\hat{x}$, which is then refined by the residual network. The final output is computed as ${x} = \mathrm{clip}(\hat{x} + R(\hat{x}), 0, 1)$.}
    \label{fig:variational_vision_pipeline}
\end{figure*}

We propose a two-stage denoising pipeline for RGB images corrupted by additive white Gaussian noise with standard deviation $\sigma=50$. In the first stage, an Attention U-Net predicted an initial clean image from the noisy input. In the second stage, a residual refinement network took the base denoised image as input and predicted a corrective residual. We obtained the final restored image by adding the predicted residual to the base output and clipping intensities to the range $[0,1]$. An overview of the pipeline is shown in \cref{fig:variational_vision_pipeline}.

The first stage was responsible for the main denoising process, while the second stage focused on correcting remaining artifacts and recovering fine image structures. We performed inference patch-wise with overlap and Gaussian blending for full-resolution restoration. We additionally applied self-ensemble test-time augmentation~\cite{timofte2016seven} to the base model.
We denote the base model output as $\hat{x}$, 
and the final restored image as $x$.

\subsubsection{Stage 1: Base Attention U-Net}

The first-stage denoiser was an Attention U-Net inspired by the U-Net architecture~\cite{ronneberger2015unet} and attention-gated skip connections~\cite{oktay2018attention}. The model followed a symmetric encoder–decoder structure with three encoder stages, one bottleneck stage, and three decoder stages.

\noindent
\textbf{Encoder}
Each encoder block contained:
\begin{itemize}
\item Two $3\times3$ convolutional layers
\item Batch Normalization after each convolution
\item GELU activation
\item Dropout (rates: 0.05, 0.05, 0.10 for levels 1--3)
\item $2\times2$ MaxPooling for downsampling
\end{itemize}

The number of feature channels increased as:
\[
64 \rightarrow 128 \rightarrow 256
\]

\noindent
\textbf{Bottleneck}
The bottleneck consisted of a convolutional block with 512 filters and dropout rate 0.15.

\noindent
\textbf{Decoder}
Each decoder block included:
\begin{itemize}
\item Upsampling by a factor of 2
\item Attention-gated skip connection
\item Concatenation with attended encoder features
\item Two $3\times3$ convolutional layers with GELU activation
\end{itemize}

The final layer was a $1\times1$ convolution with sigmoid activation producing a 3-channel RGB output. We used sigmoid activation in the final layer to constrain the output to $[0,1]$, suitable for direct image reconstruction. We chose GELU activations for the encoder and decoder blocks as they provided smoother gradients compared to ReLU, which benefited training stability. We used mixed precision training for computational efficiency, while keeping the output layer in float32 for numerical stability.

\subsubsection{Stage 2: Residual Refinement Network}

The second-stage model was a lightweight U-Net-style residual refiner. It took the first-stage output $\hat{x}$ as input and predicted a residual correction $R(\hat{x})$. We computed the final refined image as
\begin{equation*}
x = \mathrm{clip}\big(\hat{x} + R(\hat{x}), 0, 1\big).
\end{equation*}

The residual refiner used three encoder stages, one bottleneck, and three decoder stages with channel progression
\[
32 \rightarrow 64 \rightarrow 128 \rightarrow 256.
\]
Each block contained two $3\times3$ convolutional layers followed by batch normalization and ReLU activation. We used dropout rates of 0.05, 0.05, and 0.10 in the encoder and 0.15 in the bottleneck. The final $1\times1$ convolution used tanh activation to allow the network to predict both positive and negative residual corrections. During training we embedded the residual network inside a wrapper so that the clipped sum of base output and predicted residual was optimized directly against the clean target image.

\subsubsection{Dataset and Training Strategy}

We normalized all images to $[0,1]$. We added Gaussian noise with $\sigma=50/255$ to clean patches and clipped noisy inputs to $[0,1]$.

\noindent
\textbf{Base model dataset}
We trained the base Attention U-Net on the LSDIR dataset~\cite{li2023lsdir} using an image-disjoint split (80\% train / 20\% validation). For each image, we extracted eight $96\times96$ patches: one center crop, one high-texture crop selected via Laplacian-variance scoring over a $4\times4$ grid, and six random crops. We augmented training patches with random horizontal/vertical flips and $90^\circ$ rotations. Validation used five deterministic crops (four corners and center) with fixed-seed noise generation for reproducible PSNR evaluation.

\noindent
\textbf{Residual model dataset}
We trained the residual refiner using cached pairs of base-model outputs and their corresponding clean targets. We first passed noisy patches through the pretrained base model to produce $\hat{x}$ and stored both the base outputs and the corresponding clean patches as compressed \texttt{.npz} files to avoid repeated base-model inference.

We mixed residual training samples from the LSDIR dataset and the Describable Textures Dataset (DTD)~\cite{cimpoi14describing} in an 85\%/15\% ratio. The additional texture data improved the model's ability to recover high-frequency structures and complex surface patterns during residual refinement. LSDIR contributed four random $96\times96$ patches per image, while texture samples used center crops. Validation used center crops only. We followed a 90\%/10\% train--validation split for both datasets.

\noindent
\textbf{Training Strategy}
We trained the base denoiser using the L1 loss (Mean Absolute Error) with the AdamW optimizer. We chose L1 loss as it is less sensitive to outliers and encourages sharper predictions compared to MSE. For AdamW, we used default values ($\beta_1=0.9$, $\beta_2=0.999$, $\epsilon=1\times10^{-7}$) with weight decay $1\times10^{-5}$. We set the initial learning rate to $1\times10^{-4}$ with a batch size of 32 and patch size of $96\times96$. We trained for up to 50 epochs. We reduced the learning rate by a factor of 0.5 when the validation loss did not improve for 5 consecutive epochs (minimum learning rate $1\times10^{-6}$), and we applied early stopping with patience 15 based on validation PSNR. We selected the final model according to the highest validation PSNR.

We trained the second-stage residual refiner using MSE loss with the Adam optimizer. We chose MSE for the residual refiner as it penalizes larger errors more strongly, which is beneficial for fine residual corrections. We set the initial learning rate to $1\times10^{-4}$ with a batch size of 64 and patch size $96\times96$, and we trained for up to 50 epochs. The network predicted a residual correction applied to the base output, and the final result was optimized directly against the clean target image. We applied learning rate reduction on plateau and early stopping based on validation loss.

We trained both models on an NVIDIA A100 40GB SXM4 GPU rented via Lambda Labs cloud platform.

\subsubsection{Inference}
At test time, we processed fullresolution images patch-wise using $96\times96$ patches with overlap 60. We merged overlapping predictions using a Gaussian blending window to reduce visible seams and boundary artifacts.

For the base model, we applied 8-transform self-ensemble test-time augmentation~\cite{timofte2016seven}. The transform set consisted of four rotations ($0^\circ$, $90^\circ$, $180^\circ$, $270^\circ$) and their horizontally flipped counterparts. We denoised each transformed image independently, inverse-transformed back to the original orientation, and averaged to obtain the first-stage prediction.

We passed the averaged base result to the residual refinement model. The residual model predicted an additive correction, and we computed the final restored image by adding this correction to the base output and clipping intensities to $[0,1]$.
\subsection{YuFans}

Our approach ensembles two architecturally complementary denoising networks---Restormer~\cite{zamir2022restormer} and SCUNet~\cite{zhang2023practical}---both fine-tuned on the large-scale LSDIR dataset~\cite{li2023lsdir}. The key insight driving our method is that large-scale fine-tuning data is the dominant performance factor: fine-tuning on only 800 DIV2K images consistently \emph{degraded} pretrained model performance ($-$0.06 to $-$0.19\,dB), whereas fine-tuning on 85K LSDIR images yielded a +3.2\,dB improvement.

\subsubsection{Overall Pipeline}

Given a noisy input $\mathbf{I}_n \in \mathbb{R}^{H \times W \times 3}$ corrupted by additive Gaussian noise ($\sigma = 50$), we independently process it through two denoising networks and fuse their outputs:
\begin{equation}
    \hat{\mathbf{I}} = w_R \cdot f_R(\mathbf{I}_n) + w_S \cdot f_S(\mathbf{I}_n),
\end{equation}
where $f_R$ denotes the fine-tuned Restormer, $f_S$ denotes the fine-tuned SCUNet, and the ensemble weights are $w_R = 0.70$, $w_S = 0.30$. These weights were optimized via exhaustive grid search on the validation set (step size 0.05).

\subsubsection{Model Architectures}

\textbf{Restormer}~\cite{zamir2022restormer} is a multi-scale hierarchical Transformer that applies \emph{transposed} self-attention along the channel dimension, enabling efficient processing of high-resolution images without quadratic spatial complexity. We use the standard configuration: base dimension $d = 48$, encoder blocks $[4, 6, 6, 8]$, attention heads $[1, 2, 4, 8]$, FFN expansion factor $\gamma = 2.66$, and bias-free layer normalization. The network produces a residual prediction, with the final output being $\hat{\mathbf{I}} = \mathbf{I}_n + \mathbf{R}$. Total parameters: 26.1M.

\textbf{SCUNet}~\cite{zhang2023practical} combines Swin Transformer blocks with convolutional layers in a UNet topology. The hybrid design captures both long-range dependencies (via windowed self-attention) and local texture details (via convolutional branches). Configuration: base dimension 64, blocks per level $[4, 4, 4, 4, 4, 4, 4]$. Total parameters: 15.2M.

The two architectures are complementary in their error profiles: Restormer excels at global structure restoration via channel attention, while SCUNet preserves fine-grained details through its convolutional branches. Their ensemble consistently outperforms either model alone.

\subsubsection{Training Strategy}

Both models are initialized from their official pretrained Gaussian denoising ($\sigma = 50$) checkpoints and fine-tuned using a progressive training recipe:

\textbf{Restormer (two-stage progressive training):}
\begin{itemize}
    \item \textbf{Stage 1:} Patch size $256 \times 256$, batch size 4, learning rate $2 \times 10^{-5}$, cosine annealing to $10^{-7}$, 14K iterations. Val PSNR: 30.54\,dB.
    \item \textbf{Stage 2:} Resume from Stage~1 best checkpoint; patch size $512 \times 512$, batch size 1, learning rate $5 \times 10^{-6}$, cosine annealing, 20K iterations. Val PSNR: 30.57\,dB.
\end{itemize}

\textbf{SCUNet:} Patch size $256 \times 256$, batch size 4, learning rate $2 \times 10^{-5}$, cosine annealing to $10^{-7}$, 14K iterations. Val PSNR: 30.48\,dB.

\textbf{Common settings:} L1 loss; AdamW optimizer ($\beta_1 = 0.9$, $\beta_2 = 0.999$, weight decay $10^{-4}$); gradient clipping (max norm 0.01); data augmentation (horizontal/vertical flip, 90/180/270$^\circ$ rotation); noise generation follows the challenge protocol---AWGN with $\sigma = 50$ added in the $[0, 1]$ float domain, then clipped and quantized to 8-bit integers.

\subsubsection{Training Data}

We train on a combined dataset of approximately 88K high-resolution images:

\begin{center}
\begin{tabular}{lrl}
\toprule
Dataset & Images & Source \\
\midrule
LSDIR~\cite{li2023lsdir} & 84,991 & HuggingFace (parquet) \\
Flickr2K & 2,650 & Community mirror \\
DIV2K~\cite{agustsson2017ntire} (train) & 800 & Official \\
\midrule
\textbf{Total} & \textbf{88,441} & \\
\bottomrule
\end{tabular}
\end{center}

\subsubsection{Inference}

During inference, we employ the following strategies:

\textbf{Test-Time Augmentation (TTA):} We apply 8-fold geometric augmentation (4 rotations $\times$ 2 horizontal flips). Each augmented view is independently denoised, inverse-transformed, and averaged to produce the final output.

\textbf{Tiled Processing:} To handle arbitrary image sizes within GPU memory constraints, we use tiled inference with tile size 512 and overlap 256 pixels. Overlapping regions are blended using a 2D Hann window to avoid boundary artifacts.

\textbf{Ensemble Fusion:} Each model independently processes all 8 TTA views via tiled inference, and the two model outputs are fused with the optimal weights ($w_R = 0.70$, $w_S = 0.30$) before final averaging.

\subsubsection{Experimental Results}

We evaluate our method on the NTIRE 2026 Image Denoising validation and test sets:

\begin{table}[h]
\centering
\begin{tabular}{lcc}
\toprule
Method & Val PSNR & Test PSNR \\
\midrule
Restormer (pretrained, TTA) & 27.51 & --- \\
SCUNet (pretrained, TTA) & 27.48 & --- \\
\midrule
Restormer S2 (LSDIR-FT, TTA) & 30.70 & --- \\
SCUNet (LSDIR-FT, TTA) & 30.64 & --- \\
\midrule
\textbf{Ensemble (0.70\,/\,0.30)} & \textbf{30.72} & \textbf{29.80} \\
\bottomrule
\end{tabular}
\caption{Ablation results for Team YuFans. Fine-tuning on LSDIR provides +3.2\,dB over pretrained baselines.}
\end{table}

\subsubsection{Complexity}

\begin{center}
\begin{tabular}{lr}
\toprule
Metric & Value \\
\midrule
Total parameters & 41.3\,M \\
Restormer parameters & 26.1\,M \\
SCUNet parameters & 15.2\,M \\
Inference time (w/ TTA, H800) & $\sim$25\,s/image \\
Training GPU hours (total) & $\sim$20\,h \\
Hardware & 2$\times$ NVIDIA H800 80\,GB \\
\bottomrule
\end{tabular}
\end{center}

The val-to-test gap of 0.92\,dB (30.72 $\to$ 29.80) indicates notable distribution shift between the DIV2K validation images and the test set. We observed that high-overlap tiling and model weight averaging (``model soup'') improved validation PSNR but did \emph{not} transfer to test, further confirming this distribution mismatch.

\subsection{I2WM\&JNU}

\begin{figure*}[htbp]
    \centering
    \includegraphics[width=0.98\textwidth]{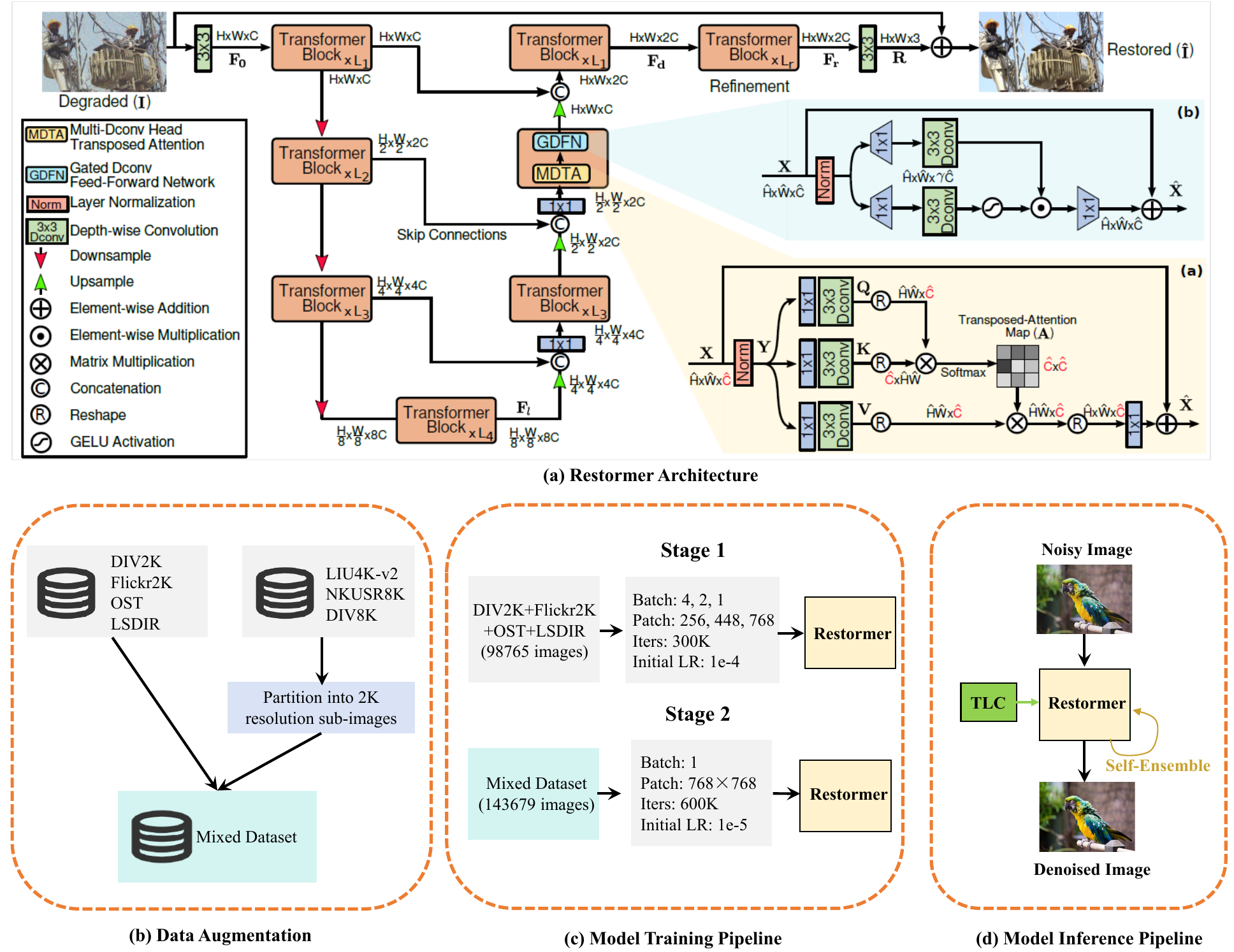}
    \caption{Overview of the proposed image denoising framework based on Restormer, incorporating large-scale training and inference-time refinement strategies (Team I2WM\&JNU).}
    \label{02pipeline}
    \vspace{-.2mm}
\end{figure*}

\textbf{Description}. The solution proposed by Team I2WM\&JNU is illustrated in \cref{02pipeline}. To achieve high-quality image denoising under limited training data and computational constraints, we adopt Restormer~\cite{zamir2022restormer}—an efficient Transformer architecture that has demonstrated strong performance in image restoration tasks—as the foundation of our approach. Our method is built upon two key pillars: large-scale training data and inference-time enhancement strategies.

Transformer-based models have consistently demonstrated strong performance in image denoising, as exemplified by Restormer. As illustrated in \cref{02pipeline}(a), Restormer incorporates two key components: the multi-Dconv head transposed attention (MDTA) block and the gated-Dconv feedforward network (GDFN). In the MDTA block, self-attention is applied across channels rather than the spatial dimension, enabling cross-channel covariance computation and generating an attention map that implicitly encodes global context. Depth-wise convolutions are applied prior to feature covariance computation to emphasize local context. In contrast, the GDFN block integrates a novel gating mechanism along with depth-wise convolutions to enhance feature transformation and information flow. Together, these designs enable Restormer to effectively capture both local and global dependencies, making it well-suited for image denoising tasks.

\textbf{Implementation Details}. To fully exploit the potential of Restormer, we construct a large-scale training set comprising high-quality images from diverse sources, including DIV2K, Flickr2K, OST, LSDIR, LIU4K-v2, NKUSR8K, and DIV8K. For LIU4K-v2, NKUSR8K, and DIV8K, the original high-resolution images are first divided into non-overlapping sub-images of approximately 2K resolution. Following this partitioning, a total of 143,679 images are obtained, providing rich and diverse training samples. As shown in \cref{02pipeline}(c), the training of Restormer is carried out in two stages.

In the first stage, the model is trained on four NVIDIA H200 GPUs using a progressive training strategy to enhance robustness and efficiency. This stage utilizes the DIV2K, Flickr2K, OST, and LSDIR datasets. The patch size is progressively increased from 256 to 448 to 768, with corresponding batch sizes of 4, 2, and 1, respectively. Optimization is performed using the AdamW optimizer for 300K iterations, with an initial learning rate of $1\times10^{-4}$ and mean squared error (MSE) loss.

In the second stage, the model is further fine-tuned on an expanded dataset that additionally incorporates LIU4K-v2, NKUSR8K, and DIV8K, also on four NVIDIA H200 GPUs. The patch size is fixed at 768 with a batch size of 4. Training continues for an additional 600K iterations using the AdamW optimizer with a reduced initial learning rate of $1\times10^{-5}$, while still optimizing with MSE loss. This large-scale training enables the model to learn more robust and generalizable features.

During inference, we further boost the performance of Restormer through two enhancement strategies. First, a self-ensemble strategy~\cite{timofte2016seven} is employed, which aggregates predictions from multiple augmented versions of the input image to produce a more stable and accurate output. Second, the Test-time Local Converter (TLC) technique is adopted to refine local details and further improve reconstruction quality. The combination of large-scale training and inference-time refinements constitutes the main contribution of our method, leading to superior denoising performance.
\subsection{Noice}
\subsubsection{General Overview}
 
\begin{figure*}[h]
  \centering
  \includegraphics[width=0.98\linewidth]{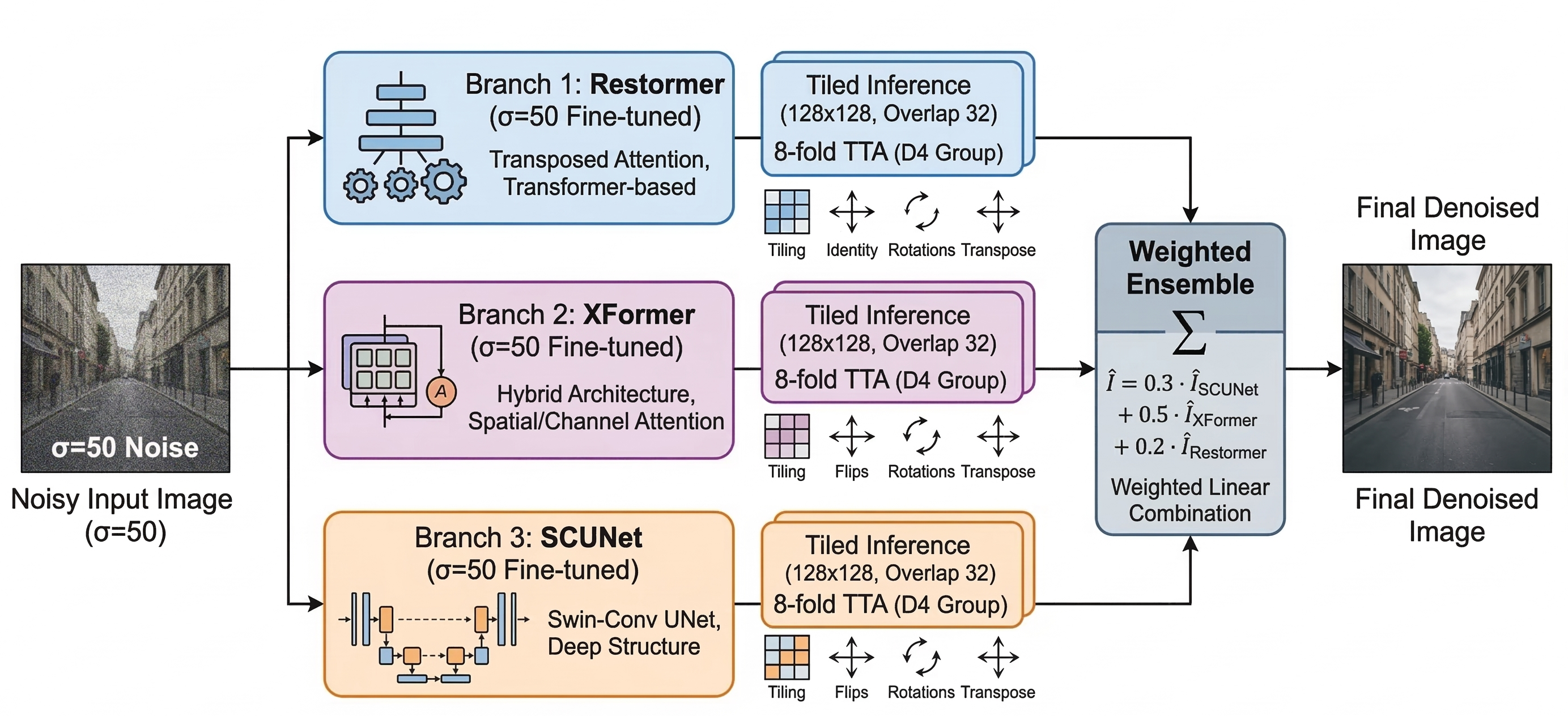}
  \caption{Overview of the \textbf{SRX} pipeline proposed by team Noice.}
  \label{fig:pipeline}
\end{figure*}

We present \textbf{SRX}, our proposed pipeline for the NTIRE 2026 Image Denoising ($\sigma=50$) challenge. SRX is built upon an ensemble of three state-of-the-art transformer- and CNN-based image restoration networks: Restormer~\cite{zamir2022restormer}, XFormer~\cite{zhang2023xformer}, and SCUNet~\cite{zhang2023practical}. Each model is independently fine-tuned on a combined high-quality dataset and subsequently fused via a weighted ensemble strategy at inference time. Test-time augmentation (TTA) with tiled inference is further employed to boost performance and handle high-resolution inputs efficiently. An overview of the SRX pipeline is illustrated in \cref{fig:pipeline}.
 
\subsubsection{Architecture Details}
 
\noindent\textbf{Restormer.}
Restormer~\cite{zamir2022restormer} is a hierarchical encoder-decoder transformer that applies multi-head self-attention in a transposed manner across channels rather than spatial dimensions, making it computationally tractable for high-resolution images. We use the standard configuration with $\texttt{dim}=48$, \texttt{num\_blocks}$=[4,6,6,8]$, \texttt{num\_refinement\_blocks}$=4$, \texttt{heads}$=[1,2,4,8]$, FFN expansion factor of $2.66$, and \texttt{BiasFree} layer normalisation (as required for the Gaussian denoising variant). The model is initialised from the official pre-trained checkpoint \texttt{gaussian\_color\_denoising\_sigma50.pth}~\cite{zamir2022restormer}.
 
\noindent\textbf{XFormer.}
XFormer~\cite{zhang2023xformer} combines spatial window-based self-attention and channel-wise attention in a unified architecture, giving complementary receptive fields to Restormer. We configure it with $\texttt{dim}=48$, \texttt{num\_blocks}$=[2,4,4]$, \texttt{spatial\_num\_blocks}$=[2,4,4,6]$, \texttt{num\_refinement\_blocks}$=4$, \texttt{heads}$=[1,2,4,8]$, and window size $16\times16$ at each scale. The model is initialised from its publicly released denoising weights and subsequently fine-tuned on our combined dataset.
 
\noindent\textbf{SCUNet.}
SCUNet~\cite{zhang2023practical} is a Swin-Conv UNet that integrates Swin Transformer blocks and convolutional residual blocks within a symmetric encoder-decoder structure. We use $\texttt{config}=[4,4,4,4,4,4,4]$ (seven-stage depth) and $\texttt{dim}=64$. The model is initialised from the official \texttt{scunet\_color\_50.pth} checkpoint~\cite{zhang2023practical} targeting $\sigma=50$ Gaussian colour noise, and then further fine-tuned.
 
\subsubsection{Training Data}
 
All three models are fine-tuned on a combined dataset of high-quality clean images:
\begin{itemize}
  \item \textbf{LSDIR}~\cite{li2023lsdir}: 2{,}000 images sampled from the Large Scale Dataset for Image Restoration.
  \item \textbf{DIV2K}~\cite{agustsson2017ntire}: 800 high-resolution training images from the DIV2K dataset.
\end{itemize}
 
Gaussian noise with $\sigma=50$ is synthesised on-the-fly during training, following standard self-supervised Gaussian denoising protocols. No external noisy images are used; ground-truth clean patches serve as targets. No additional data beyond the above two datasets is used.
 
\subsubsection{Data Augmentation}
 
During training, the following augmentation pipeline is applied to each sampled patch:
\begin{itemize}
  \item \textbf{Random crop:} Patches of size $64\times64$ are randomly cropped from each image. If the image is smaller than the crop size, it is upscaled to $64\times64$ via bilinear interpolation.
  \item \textbf{On-the-fly noise synthesis:} Additive white Gaussian noise with $\sigma=50$ is added to each clean crop, producing the corresponding noisy input.
  \item \textbf{Horizontal and vertical flips:} Random left-right and top-bottom flips are applied.
  \item \textbf{Random rotations:} Random $90^{\circ}$ rotations are applied for orientation diversity.
\end{itemize}
 
\subsubsection{Training Strategy}
 
Each model is fine-tuned independently using the following protocol.
 
\noindent\textbf{Loss function.} We use the $\ell_1$ (mean absolute error) loss between the network output and the clean ground-truth patch:
\begin{equation}
  \mathcal{L} = \frac{1}{N}\sum_{i=1}^{N} \|\hat{x}_i - x_i\|_1,
\end{equation}
where $\hat{x}_i$ is the restored patch and $x_i$ is the clean reference.
 
\noindent\textbf{Optimiser.} AdamW~\cite{adamw} with an initial learning rate of $1\times10^{-4}$ and default momentum parameters ($\beta_1=0.9$, $\beta_2=0.999$).
 
\noindent\textbf{Batch and patch size.} Batch size of 4 and patch size of $64\times64$.
 
\noindent\textbf{Epochs.} 40 epochs for each model.
 
\noindent\textbf{Mixed precision.} Training uses PyTorch Automatic Mixed Precision (AMP) with a \texttt{GradScaler} to reduce GPU memory consumption and accelerate training.
 
\noindent\textbf{Partial freezing (Restormer only).} For Restormer, the three encoder levels (\texttt{encoder\_level1}, \texttt{encoder\_level2}, \texttt{encoder\_level3}) are frozen during fine-tuning. Only the bottleneck, decoder, and refinement blocks are updated. This prevents overfitting the low-level feature extraction layers that are already well trained from the original pre-training, while allowing the higher-level reconstruction components to adapt to the challenge data distribution.
 
\subsubsection{Inference: Tiled Processing}
 
Because the images may be arbitrarily large, direct full-image inference is impractical. We employ overlapping tile-based inference:
\begin{itemize}
  \item \textbf{Tile size:} $128\times128$ pixels.
  \item \textbf{Overlap:} 32 pixels on each border (stride = $128 - 32 = 96$).
  \item Overlapping regions are accumulated and averaged by a pixel-wise weight map, eliminating seam artefacts at tile boundaries.
\end{itemize}
 
\subsubsection{Test-Time Augmentation (TTA)~\cite{timofte2016seven}}
 
To further improve prediction quality, we apply an 8-fold geometric TTA covering all symmetries of the dihedral group $D_4$:
\begin{enumerate}
  \item Original (identity)
  \item Horizontal flip
  \item Vertical flip
  \item Horizontal + Vertical flip ($180^{\circ}$ rotation)
  \item Transpose
  \item Transpose + Horizontal flip
  \item Transpose + Vertical flip
  \item Transpose + Horizontal + Vertical flip
\end{enumerate}
Each augmented version of the input is passed through the model (with tiled inference), the inverse geometric transformation is applied to the output, and all eight results are averaged pixel-wise.
 
\subsubsection{Ensemble}
 
The final denoised image is produced by a weighted linear combination of the outputs of all three fine-tuned models after TTA:
\begin{equation}
  \hat{I} = 0.3 \cdot \hat{I}_{\text{SCUNet}} + 0.5 \cdot \hat{I}_{\text{XFormer}} + 0.2 \cdot \hat{I}_{\text{Restormer}},
\end{equation}
where $\hat{I}_{\text{SCUNet}}$, $\hat{I}_{\text{XFormer}}$, and $\hat{I}_{\text{Restormer}}$ are the TTA-averaged outputs of SCUNet, XFormer, and Restormer, respectively. The weights were determined empirically on the CodaBench validation leaderboard. XFormer receives the highest weight as it demonstrated the strongest individual performance on the validation split.
 
\subsection{Titans}

Our approach is based on the UnifyFormer architecture, which investigates the integration of group-level spatial reasoning with channel attention for image denoising. The goal of the network is to restore clean images from noisy observations corrupted with additive Gaussian noise.

\subsubsection{Overall Pipeline}

\begin{figure*}[t]
\begin{center}
\scalebox{1}{
    \begin{tabular}[t]{c} \hspace{-2mm}
    
    \includegraphics[width=\textwidth]{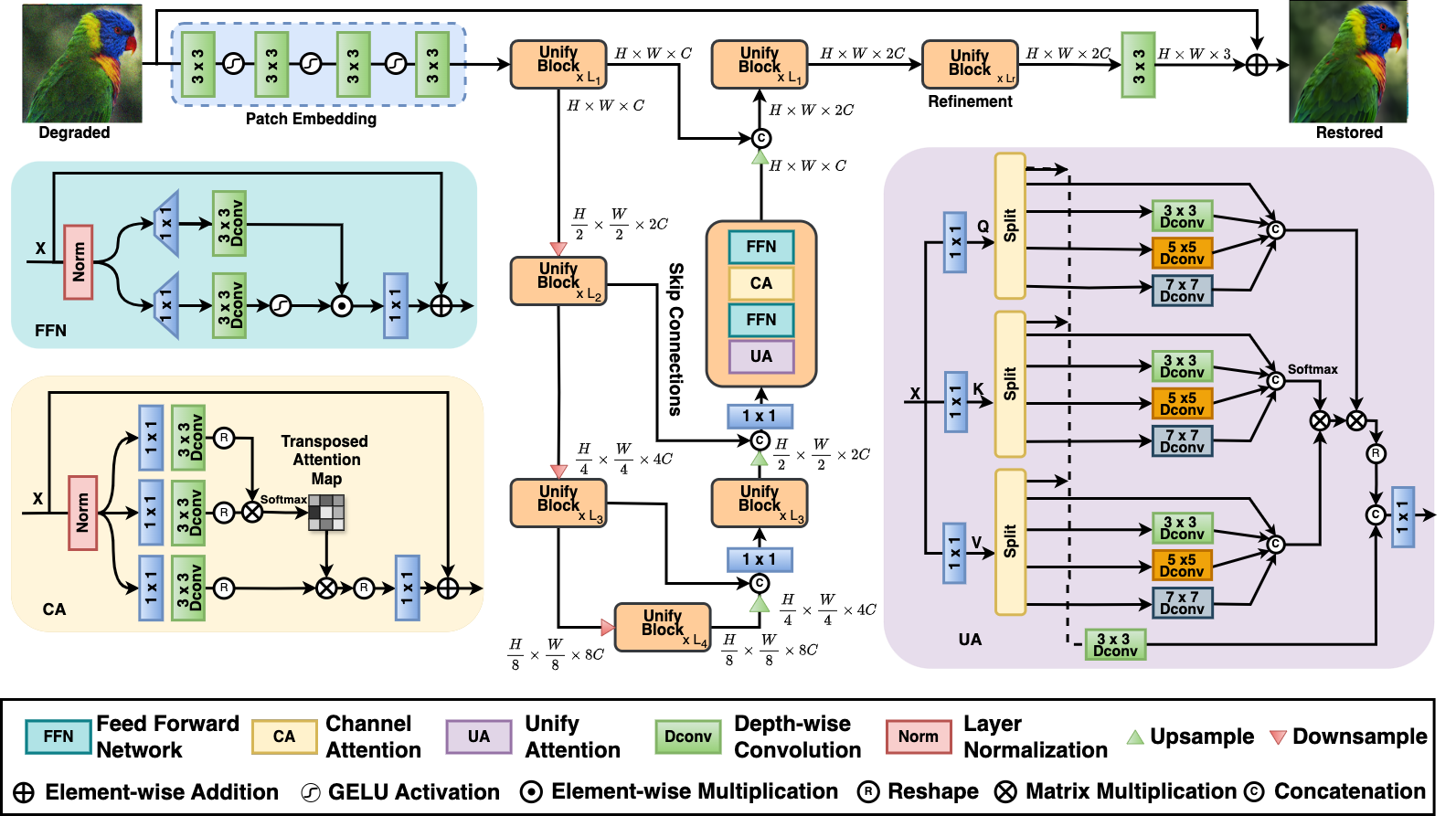}
    \end{tabular}
    }
\end{center}
\vspace{-0.7cm}
\caption{Overall framework of UnifyFormer proposed by team Titans.}
\label{fig:framework}
\end{figure*}

We present \textbf{UnifyFormer} (\cref{fig:framework}), a transformer-based architecture for image 
restoration under Gaussian noise ($\sigma = 50$). The network operates on a 
degraded input $\mathbf{I} \in \mathbb{R}^{H \times W \times 3}$, passing it 
through a multi-stage patch embedding to extract low-level features, which are 
then processed through a four-level hierarchical encoder-decoder built from 
our proposed \textbf{UnifyBlocks}. Skip connections between encoder and decoder 
levels are fused via concatenation followed by $1{\times}1$ convolution. A 
refinement stage at the decoder output further sharpens the recovered features. 
The network produces a residual image $\mathbf{R}$, and the final restored 
output is $\hat{\mathbf{I}} = \mathbf{I} + \mathbf{R}$.

\subsubsection{Patch Embedding}

Instead of using a single shallow convolution, we adopt a progressive patch 
embedding module composed of four successive $3{\times}3$ convolutions. Let 
$d$ denote the base feature dimension. The embedding module expands the input 
channels as
\[
3 \rightarrow d/4 \rightarrow d/2 \rightarrow d/2 \rightarrow d,
\]
with GELU activations after the first three convolutions. This design injects 
strong local inductive bias at the earliest stage and provides richer low-level 
features before attention-based processing.

\subsubsection{UnifyBlock and Unify Attention (UA)}

The central contribution of UnifyFormer is \textbf{Unify Attention}, which augments 
self-attention with explicit multi-scale local aggregation. Given an input feature 
map $\mathbf{X} \in \mathbb{R}^{B \times d \times H \times W}$, we first obtain 
queries, keys, and values through a $1{\times}1$ convolution:
\begin{equation}
    [\mathbf{Q}, \mathbf{K}, \mathbf{V}] = \phi(\mathbf{X}),
\end{equation}
where $\phi(\cdot)$ denotes the linear projection.

Each of $\mathbf{Q}$, $\mathbf{K}$, and $\mathbf{V}$ is then partitioned into 
five equal channel groups of size $d_s = d/5$. Four groups are independently 
processed by separable convolutions with kernel sizes $\{1,3,5,7\}$:
\begin{equation}
    \mathbf{Z}^{(i)} = \delta_i\!\left(\mathrm{SepConv}_{k_i}(\mathbf{X}^{(i)})\right),
    \quad k_i \in \{1,3,5,7\},
\end{equation}
where $\delta_i$ denotes GELU activation. These branches capture local context 
at multiple receptive fields.

The fifth group is not treated as an independent spatial branch. Instead, the 
fifth groups from $\mathbf{Q}$, $\mathbf{K}$, and $\mathbf{V}$ are concatenated 
and fused to produce a compact local proxy:
\begin{equation}
    \mathbf{X}_{\mathrm{agg}} =
    \psi\!\left(
    \mathrm{Concat}\big(
    \mathbf{Q}^{(5)}, \mathbf{K}^{(5)}, \mathbf{V}^{(5)}
    \big)\right),
\end{equation}
where $\psi(\cdot)$ denotes a separable $3{\times}3$ convolution followed by 
layer normalization and GELU. This fused representation encodes local structure 
shared across the projected query, key, and value streams. The remaining four 
processed groups are concatenated to form the attention features.

To reduce complexity, UA uses a linearized attention formulation. After reshaping 
into multi-head form, attention is computed as
\begin{equation}
    \tilde{\mathbf{K}} = \mathrm{softmax}_{n}(\mathbf{K}), \qquad
    \mathbf{O}_{\mathrm{UA}} =
    \alpha\, \mathbf{Q}\big(\tilde{\mathbf{K}}^{\top}\mathbf{V}\big),
\end{equation}
where the softmax is applied along the spatial-token dimension $n$, and 
$\alpha = d_h^{-1/2}$ is the standard head-wise scaling factor with $d_h$ the 
feature dimension per head.

Finally, the local proxy is concatenated back with the attention output to restore 
the original channel dimension:
\begin{equation}
    \mathbf{O} =
    \mathrm{Proj}\!\left(
    \mathrm{Concat}(\mathbf{O}_{\mathrm{UA}}, \mathbf{X}_{\mathrm{agg}})
    \right),
\end{equation}
where $\mathrm{Proj}(\cdot)$ is a final $1{\times}1$ convolution. In this way, 
UA combines efficient global interaction with explicit multi-scale local context.

\subsubsection{Channel Attention (CA)}

We complement UA with the channel attention module from Restormer~\cite{restormer}, 
which models inter-channel dependencies with linear complexity. Given an input 
feature map, queries, keys, and values are first produced through a $1{\times}1$ 
convolution followed by a depth-wise $3{\times}3$ convolution. They are then 
reshaped into $\mathbb{R}^{B \times h \times c \times HW}$, where $h$ denotes 
the number of heads and $c$ the channels per head. After $\ell_2$ normalization 
of queries and keys, the channel attention map is computed as
\begin{equation}
    \mathbf{A}_c =
    \mathrm{softmax}\!\left(
    \hat{\mathbf{Q}}\hat{\mathbf{K}}^{\top}\cdot \tau
    \right), \qquad
    \mathbf{O}_c = \mathbf{A}_c \mathbf{V},
\end{equation}
where $\tau$ is a learnable temperature parameter for each attention head. Since 
the similarity is computed across channels rather than spatial positions, CA is 
less sensitive to localized noise corruption and provides stable global channel 
recalibration.

\subsubsection{Feed-Forward Network}

Each attention module is followed by a gated feed-forward network. Given input 
features $\mathbf{F}$, the FFN first expands the channel dimension using a 
$1{\times}1$ convolution, applies a depth-wise $3{\times}3$ convolution, splits 
the result into two branches, and performs gated interaction:
\begin{equation}
    \mathbf{Z} = \mathrm{DWConv}(\mathbf{W}_{\mathrm{in}}\mathbf{F}),
\end{equation}
\begin{equation}
    [\mathbf{Z}_1,\mathbf{Z}_2] = \mathrm{Split}(\mathbf{Z}),
\end{equation}
\begin{equation}
    \mathrm{FFN}(\mathbf{F}) =
    \mathbf{W}_{\mathrm{out}}\big(\mathrm{GELU}(\mathbf{Z}_1)\odot \mathbf{Z}_2\big),
\end{equation}
where $\odot$ denotes element-wise multiplication. This gating mechanism 
selectively enhances informative responses and helps recover fine image details.

\subsubsection{TransformerBlock}

Each TransformerBlock applies UA and CA sequentially, with normalization and 
residual learning at every stage:
\begin{align}
    \mathbf{x} &\leftarrow \mathbf{x} +
    \mathrm{UA}\big(\mathrm{Norm}_1(\mathbf{x})\big), \\
    \mathbf{x} &\leftarrow \mathbf{x} +
    \mathrm{FFN}\big(\mathrm{Norm}_2(\mathbf{x})\big), \\
    \mathbf{x} &\leftarrow \mathbf{x} +
    \mathrm{CA}\big(\mathrm{Norm}_3(\mathbf{x})\big), \\
    \mathbf{x} &\leftarrow \mathbf{x} +
    \mathrm{FFN}\big(\mathrm{Norm}_4(\mathbf{x})\big).
\end{align}
This ordering first enriches the representation with multi-scale spatial reasoning, 
then refines channel dependencies, with each attention stage followed by a gated 
nonlinear transformation.

\subsubsection{Training Strategy}

We perform progressive learning where the network is trained on smaller image patches in the early epochs and on gradually larger patches in later epochs. The model trained on mixed-size patches via progressive learning shows enhanced performance at test time where images can be of different resolutions (a common case in image restoration). Training is performed using images from the LSDIR \cite{lilsdir}  dataset. Synthetic Gaussian noise with noise level $\sigma = 50$ is added to generate noisy inputs.
From level-1 to level-4, the number of Transformer blocks are [2, 3, 3, 4], 
attention heads in UA and CA are [1, 2, 4, 8], and feature dimensions across 
encoder levels are [50, 100, 200, 400]. The refinement stage contains 2 blocks. 
The channel expansion factor in the FFN is $\gamma = 2.66$.

The model is trained using the AdamW optimizer with parameters 
$\beta_1 = 0.9$, $\beta_2 = 0.999$, and weight decay $1\times10^{-4}$. 
The loss function used is $\ell_1$. Training is performed for 300K iterations 
with an initial learning rate of $3\times10^{-4}$ reduced to $1\times10^{-6}$ 
using cosine annealing.

For progressive learning the patch sizes and batch sizes are scheduled as follows:

\begin{center}
\begin{tabular}{ccc}
\toprule
Patch Size & Batch Size & Iterations \\
\midrule
256 & 6 & 92K \\
288 & 4 & 64K \\
320 & 4 & 48K \\
384 & 2 & 36K \\
448 & 2 & 36K \\
512 & 1 & 24K \\
\bottomrule
\end{tabular}
\end{center}

\subsubsection{Experimental Results}

We evaluate our model on the NTIRE 2026 Image Denoising test set.

\begin{table}[!h]
\centering
\begin{tabular}{lc}
\toprule
Dataset & PSNR (dB) \\
\midrule
Test & 29.87 \\
\bottomrule
\end{tabular}
\caption{Quantitative results for image denoising ($\sigma=50$).}
\end{table}

\subsection{tarrya}

\begin{figure*}[t]
    \begin{subfigure}{0.33\textwidth}
        \includegraphics[width=\textwidth]{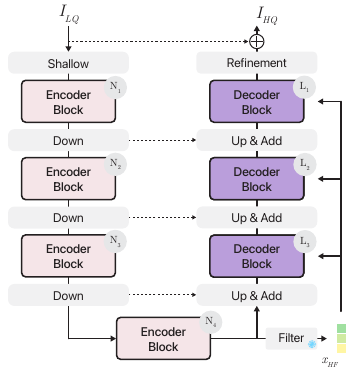}
        \subcaption{\textit{Architecture overview}}
        \label{fig:method:overview}
    \end{subfigure}
    \hfill
    \begin{subfigure}{0.33\textwidth}
        \includegraphics[width=\textwidth]{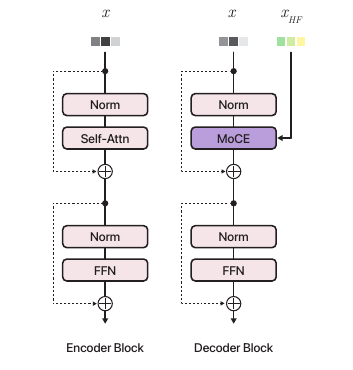}
        \subcaption{\textit{Transformer blocks.}}
        \label{fig:method:blocks}
    \end{subfigure}
    \hfill
    \begin{subfigure}{0.33\textwidth}
        \includegraphics[width=\textwidth]{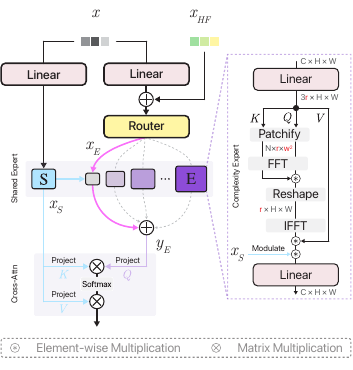}
        \subcaption{\textit{Mixture-of-complexity-experts}}
        \label{fig:method:moce}
    \end{subfigure}
    \vspace{-5mm}
    \caption{The MoCE-IR framework proposed by team tarrya. Despite recent advances in MoE-based image restoration, inconsistent expert behavior—where some experts over-generalize while others underperform—limits their computational efficiency. We address this through \textit{complexity experts}: flexible blocks with varying computational capacity and receptive fields. Our MoCE-IR employs an asymmetric encoder-decoder architecture where each decoder block contains a mixture-of-complexity-experts layer for adaptive capacity routing.}
    \label{fig:method:main}
    \end{figure*}

    \textbf{General method description:} 
    For the NTIRE Image Denoising Challenge (Noise Level = 50), we propose MoCE-IR, an image restoration network based on the Mixture-of-Complexity-Experts (MoCE) framework \cite{zamfir2025complexity}. Traditional Mixture-of-Experts (MoE) models typically employ uniform and rigid expert blocks, which limits the model's ability to bypass irrelevant experts and maximize computational efficiency during inference. To address this limitation, we introduce ``Complexity Experts'' into our network---flexible expert blocks with varying computational capacities and receptive field sizes. Since the complexity of noise degradation in the input image is not known \textit{a priori}, routing features to the appropriate experts poses a significant challenge. To this end, we design a complexity-aware task allocation strategy inspired by spring mechanics. This strategy introduces a bias that favors low-complexity experts, preferentially routing tasks to simpler experts whenever possible. This not only vastly improves computational efficiency but also spontaneously induces task-discriminative learning, allowing the model to automatically allocate tasks to the most suitable experts based on the denoising difficulty. The pipeline of the method in shown in \cref{fig:method:main}.

    \textbf{Training strategy:} 
    The model was trained on the large-scale datasets(BSD400 \cite{arjomand2017deep} and WED \cite{ma2016waterloo}). We optimized the network using the Adam optimizer with an initial learning rate of [e.g., $2 \times 10^{-4}$], gradually decreased using a Cosine Annealing strategy. The network was primarily supervised using [e.g., L1 Loss / Charbonnier Loss]. To prevent overfitting, high-resolution images and their corresponding noisy inputs were randomly cropped into [e.g., $128 \times 128$] pixel patches, with data augmentation techniques including random horizontal flips and rotations applied. The model was trained on [e.g., 8 $\times$ NVIDIA A100 GPUs] with a total batch size of [e.g., 32] for [e.g., 300,000] iterations.

    \textbf{Experimental results:} 
    Evaluated on the provided NTIRE testing set, MoCE-IR demonstrated outstanding denoising performance under the severe noise level 50 setting, achieving a PSNR of 29 dB and an SSIM of 0.84.

\subsection{APRIL-AIGC}
\subsubsection{Method Overview}
\begin{figure*}[t]
\centering
\includegraphics[width=\linewidth]{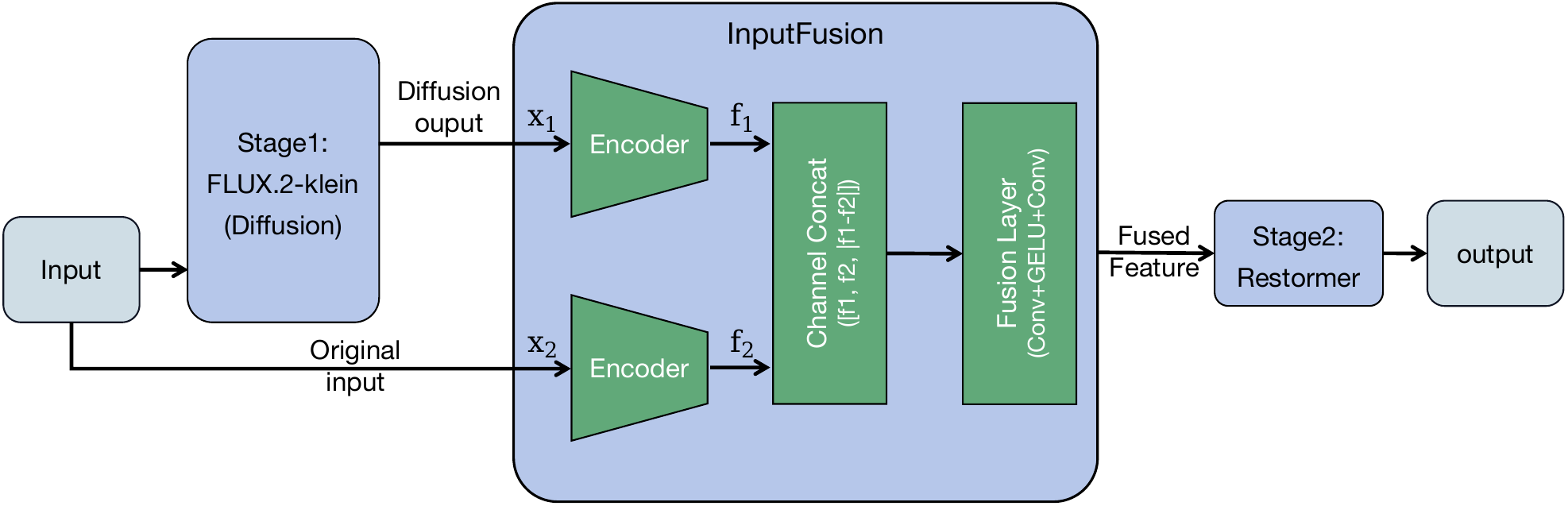}
\caption{Overview of the two-stage denoising pipeline proposed by team APRIL-AIGC. Stage 1 utilizes FLUX.2-klein-4B-base for fundamental restoration, while Stage 2 refines residual artifacts using a Restormer-based module.}
\label{fig:april_denoising_pipeline}
\end{figure*}

We present a two-stage pipeline designed for image denoising at a noise level of $\sigma=50$. Stage 1 leverages the FLUX.2-klein-4B-base model \cite{flux-2-2025}, which utilizes a Multimodal Diffusion Transformer (MM-DiT) architecture to perform the primary restoration. Stage 2 employs a Restormer-based refiner \cite{zamir2022restormer} to eliminate residual color shifts and local artifacts left by the diffusion process. An overview of the full pipeline is shown in \cref{fig:april_denoising_pipeline}.

The noisy input image is first encoded into a condition latent $z_n$ via VAE. During the denoising process, the current latent state $z_t$ and the condition latent $z_n$ are patchified, packed, and concatenated to form the transformer input:
\[
H = [\mathcal{P}(z_t); \mathcal{P}(z_n)],
\]
where $\mathcal{P}(\cdot)$ represents the latent patchification and token packing operations. A fixed denoising prompt is applied consistently across both training and inference.

In Stage 2, the refiner receives the Stage 1 output $x_1$ and the original noisy image $x_2$. We extract shallow features from both inputs and concatenate them along with their absolute difference:
\[
F = \phi([f_1, f_2, |f_1 - f_2|]),
\]
where $f_1$ and $f_2$ denote the shallow features of $x_1$ and $x_2$, respectively. The Restormer backbone consists of four levels with a base embedding dimension 48, stage depths $[4,6,6,8]$, attention heads $[1,2,4,8]$, and FFN expansion factor is $2.66$. The network predicts a residual correction:
\[
\hat{I} = x_1 + R(x_2, x_1),
\]
where $R(\cdot)$ denotes the fusion and Restormer module.

\subsubsection{Training and Inference}
Stage 1 is trained on the DIV2K \cite{agustsson2017ntire} and LSDIR \cite{li2023lsdir} datasets using random $768 \times 768$ crops. Training is conducted for 10,000 iterations on 8 \(\times\) NVIDIA H20 GPUs using the AdamW optimizer with a batch size of 4 and a learning rate of $1 \times 10^{-5}$. We apply a condition dropout probability of 0.15. After Stage 1 fully converges, we generate 2,000 paired training samples from its outputs to train the Stage 2 refiner. The refiner is then trained for 40 epochs at the same resolution using AdamW with a learning rate of $2 \times 10^{-4}$.

During inference, Stage 1 employs 5 sampling steps with a guidance scale of 2.0 and an empty string negative prompt. Full-resolution images are processed via a sliding-window tiling strategy on $768 \times 768$ patches with a 64-pixel overlap.

\textbf{Loss Functions.} The diffusion model is optimized using a rectified-flow objective \cite{liu2022flowstraightfastlearning}. Given a clean latent $z_0$ and Gaussian noise $\varepsilon \sim \mathcal{N}(0, I)$, the forward process is defined as:
\[
z_t = (1 - t)z_0 + t\varepsilon,
\]
with the target velocity:
\[
u_t = \varepsilon - z_0.
\]
The network prediction $v_t$ is optimized via the loss:
\[
L_{\mathrm{RF}} = \|v_t - u_t\|_2^2.
\]
The refiner is optimized using the Charbonnier loss \cite{charbonnier1994two}:
\[
L_{\mathrm{Charb}} = \sqrt{(\hat{I} - I_c)^2 + \delta^2},
\]
where $\hat{I}$ is the refined output, $I_c$ is the clean ground truth, and $\delta$ is a small constant for numerical stability.

On our internal validation split, the diffusion-only stage achieves 23.96 dB PSNR, 0.681 SSIM, and 0.262 LPIPS. The Stage 2 refiner further improves performance to 25.23 dB PSNR, 0.722 SSIM, and 0.246 LPIPS.

\subsection{KLETech-CEVI}

\begin{figure*}[t]
\centering
\includegraphics[width=\linewidth]{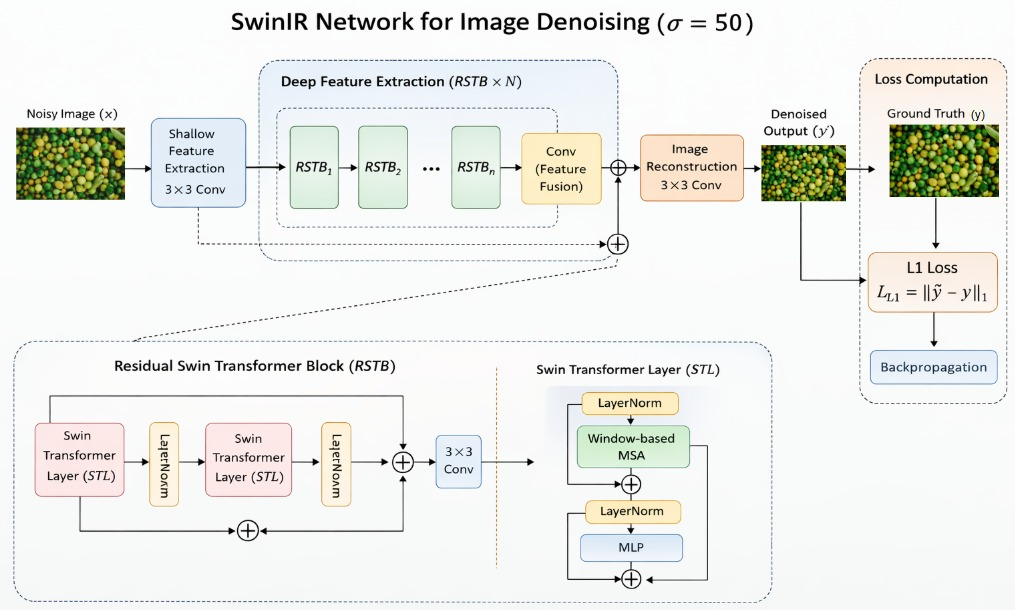}
\caption{Overall pipeline proposed by team KLETech-CEVI.}
\label{fig:team12_pipeline.png}
\end{figure*}

Our proposed solution is based on the SwinIR architecture~\cite{liang2021swinir} implemented within the BasicSR framework~\cite{wang2021basicsr}. The overall architecture of the proposed model is illustrated in \cref{fig:team12_pipeline.png}. SwinIR is a transformer-based image restoration model that utilizes shifted window multi-head self-attention to capture both local and global contextual information efficiently. The network consists of three main stages: shallow feature extraction, deep feature extraction using Swin Transformer blocks, and image reconstruction.
Given an input noisy image $x$, the shallow feature extraction layer first maps the input image into a high-dimensional feature space using a convolution layer:
\begin{equation}
F_0 = H_{SF}(x)
\end{equation}
where $F_0$ represents the shallow features and $H_{SF}$ denotes the shallow feature extraction module.
The extracted features are then processed through multiple Swin Transformer blocks which capture long-range dependencies using window-based self-attention:
\begin{equation}
F_d = H_{STB}(F_0)
\end{equation}
where $F_d$ represents the deep features and $H_{STB}$ denotes the stacked Swin Transformer blocks.
Finally, a reconstruction module generates the restored image $\hat{y}$:
\begin{equation}
\hat{y} = H_{rec}(F_d)
\end{equation}
where $H_{rec}$ denotes the reconstruction layer responsible for producing the final denoised image.

\cref{fig:team12_pipeline.png} illustrates the overall architecture of the proposed SwinIR-based denoising framework. The model takes a noisy image as input, extracts shallow features using a convolution layer, and processes them through multiple RSTB for deep feature extraction. The extracted features are fused and passed through the reconstruction module to generate the final denoised output image.

The model is trained using the dataset provided in the NTIRE 2026 Image Denoising Challenge. During training, images are randomly cropped into patches of size $128 \times 128$ to increase the number of training samples. Data augmentation techniques such as horizontal flipping and rotation are applied to improve generalization.
The network is optimized using the Adam optimizer with an initial learning rate of $2 \times 10^{-4}$ and weight decay of $1 \times 10^{-4}$. A MultiStep learning rate scheduler is used to gradually reduce the learning rate during training. The model is trained for 100,000 iterations with a batch size of 4 on single NVIDIA RTX3090.
The training objective is the L1 pixel loss, which minimizes the difference between the predicted denoised image $\hat{y}$ and the ground truth clean image $y$:
\begin{equation}
L_{L1} = \| \hat{y} - y \|_1
\end{equation}
\subsection{SNUC}

\subsubsection{Network Architecture}

The proposed approach is built upon the Restormer architecture~\cite{zamir2022restormer}, a transformer-based framework specifically designed for high-resolution image restoration. The model employs Multi-DConv Head Transposed Attention (MDTA) modules to efficiently capture long-range contextual dependencies while maintaining computational tractability. The overall structure follows a hierarchical encoder-decoder paradigm augmented with skip connections to preserve spatial detail across resolution levels.

The network is configured with an input and output channel dimensionality of 3, a base feature dimension of 48, and transformer block depths of $[4, 6, 6, 8]$ across the four encoder-decoder levels. Multi-head attention is applied with head counts of $[1, 2, 4, 8]$ at corresponding levels. The refinement stage comprises 4 transformer blocks, and the feed-forward network employs an expansion ratio of 2.66.

In this work, the Restormer architecture is adapted for high-noise image restoration through fine-tuning on the DIV2K dataset. A patch-based training strategy is employed to enable effective learning under severe noise conditions ($\sigma = 50$), demonstrating the robustness of transformer-based restoration models in challenging degradation scenarios.

\subsubsection{Training Pipeline}

The training procedure follows a patch-based supervised denoising protocol:

\begin{enumerate}
    \item Clean reference images are sourced from the DIV2K dataset.
    \item Additive white Gaussian noise with standard deviation $\sigma = 50$ is synthetically applied to each clean image.
    \item Random $128 \times 128$ crops are extracted from the corrupted images to form training patches.
    \item Each noisy patch is passed through the Restormer network to obtain the predicted clean reconstruction.
    \item The reconstruction loss is computed between the network output and the corresponding clean patch using the L1 criterion.
\end{enumerate}

\subsubsection{Training Configuration}

The model is trained on the DIV2K dataset, comprising 800 high-resolution images. Gaussian noise at $\sigma = 50$ is applied during training. Patches of size $128 \times 128$ are extracted with a batch size of 2. Optimization is performed using AdamW with a learning rate of $1 \times 10^{-5}$ and weight decay of $1 \times 10^{-4}$. The network is trained for 50 epochs using the L1 loss function.

\subsubsection{Training Results}

The model achieves its best validation performance at epoch 34, recording a peak PSNR of 32.10~dB and a minimum training loss of approximately 0.0215.

\subsubsection{Leaderboard Results}

On the competition leaderboard, the proposed method attains a PSNR of 30.4~dB on the development split and 29.90~dB on the test split, achieving a final rank of 9th. 

\subsubsection{Evaluation Metric}

Model performance is evaluated using Peak Signal-to-Noise Ratio (PSNR), defined as:

\begin{equation}
    \text{PSNR} = 10 \log_{10} \left( \frac{\text{MAX}^2}{\text{MSE}} \right)
\end{equation}

\noindent where $\text{MAX}$ denotes the maximum possible pixel intensity and $\text{MSE}$ is the mean squared error between the reconstructed and reference images.

\subsection{CV\_SVNIT}













\textbf{General Method Description.}
To address the Gaussian image denoising problem with noise level $\sigma=50$, we adopt a Transformer-based image restoration framework built upon the Restormer architecture \cite{zamir2022restormer}. Restormer has demonstrated strong performance on various image restoration tasks due to its ability to capture long-range dependencies using self-attention mechanisms.

\begin{figure*}[ht]
\centering
\setlength{\abovecaptionskip}{3pt}
\setlength{\belowcaptionskip}{-5pt}
\includegraphics[width=0.8\linewidth]{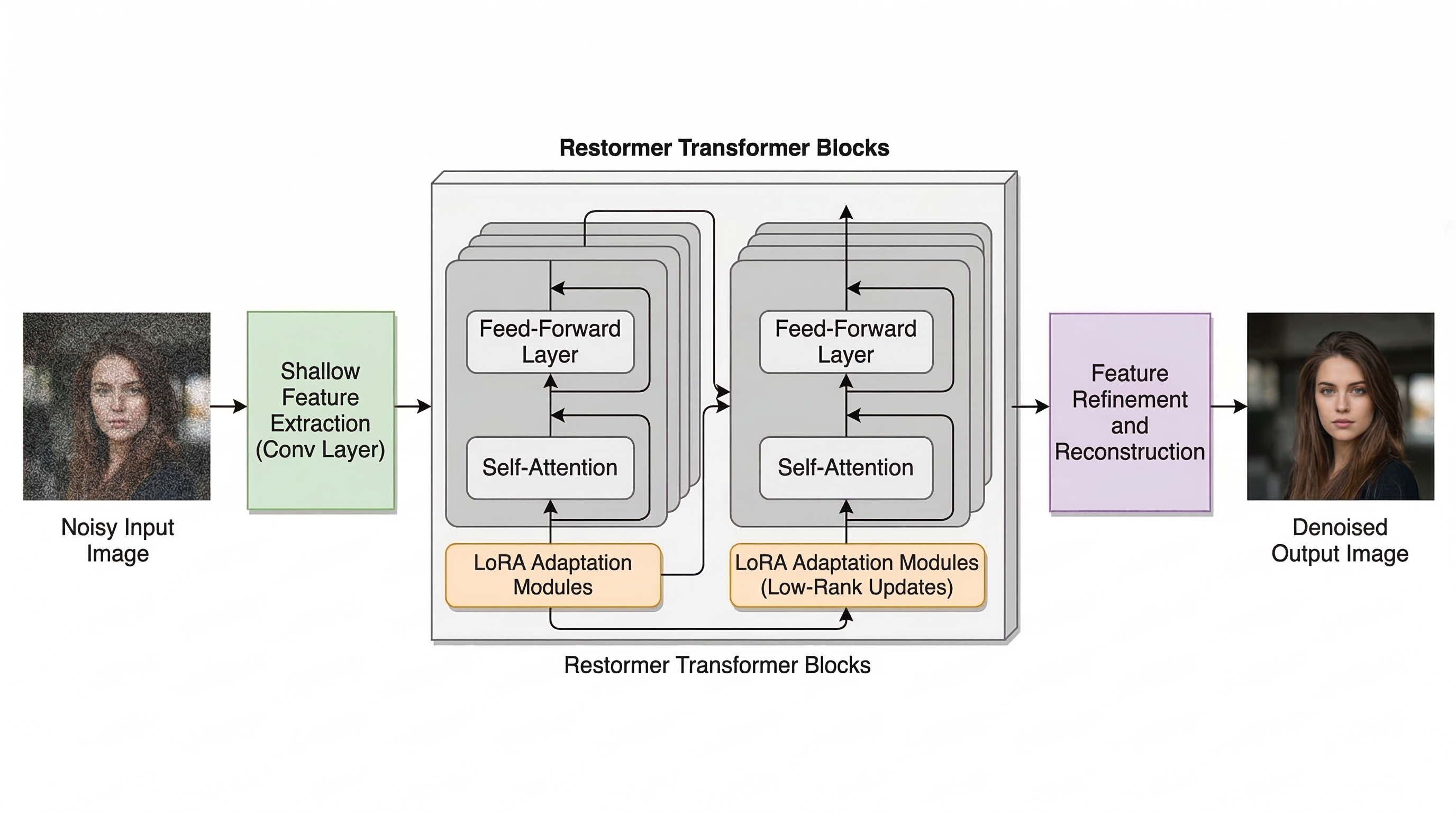}
\vspace{-32pt}
\caption{Overview of the Restormer + LoRA architecture proposed by team CV\_SVNIT.}
\label{fig:restormer_lora}
\end{figure*}

In our approach, we utilize a pretrained Restormer model and adapt it to the denoising task using Low-Rank Adaptation (LoRA) \cite{hu2022lora}. Instead of updating all network parameters, LoRA introduces trainable low-rank matrices into the attention projection layers while keeping the original pretrained weights frozen. This significantly reduces the number of trainable parameters and improves training efficiency while maintaining strong restoration performance.

\vspace{0.3em}
\textbf{Model Architecture Description.}
The proposed architecture is built upon the Restormer encoder--decoder framework. As illustrated in \cref{fig:restormer_lora}, the noisy input image is first processed through a shallow feature extraction module implemented using a convolutional layer to capture low-level spatial information.

The extracted shallow features are subsequently processed by a hierarchy of Restormer transformer blocks that perform deep feature extraction using multi-head self-attention and feed-forward layers. These attention mechanisms enable the network to capture both local image structures and long-range contextual dependencies, which are crucial for accurate image restoration.

To efficiently adapt the pretrained Restormer model to the denoising task, Low-Rank Adaptation (LoRA) modules are introduced into the attention projection layers of the transformer blocks. Each LoRA module introduces a pair of low-rank matrices that approximate weight updates while keeping the original pretrained parameters frozen. This parameter-efficient adaptation significantly reduces the number of trainable parameters while preserving the strong representational capability of the backbone network.

Finally, the refined feature representations are passed through reconstruction layers that progressively map the deep features back to the RGB image space. The reconstruction module refines spatial details and produces the final denoised output image.

\vspace{0.3em}
\textbf{Training Details.}
The model is trained using a combination of the DIV2K and LSDIR datasets. To balance dataset diversity and size, a weighted sampling strategy is used during training, where approximately 12\% of samples are drawn from DIV2K and 88\% from LSDIR.

During training, random patches of size $256 \times 256$ are cropped from the high-resolution images. Gaussian noise with $\sigma = 50$ is added on-the-fly to generate the noisy inputs.

The network is trained for $2 \times 10^4$ iterations with a batch size of 1. The training procedure is implemented using the PyTorch framework. We adopt the Charbonnier loss function, which is widely used for image restoration tasks due to its robustness and stability.

The optimizer used is AdamW with an initial learning rate of $1 \times 10^{-4}$ and weight decay of $1 \times 10^{-4}$. The learning rate is scheduled using cosine annealing to ensure stable convergence during training.

During inference, tiled processing is applied to handle high-resolution images efficiently. In addition, 8-fold test-time augmentation is employed to further improve denoising performance by averaging predictions from multiple geometric transformations of the input image.
\subsection{ML\_SVNIT}













\textbf{General Method Description.}
To address the Gaussian image denoising problem with noise level $\sigma=50$, we propose a model ensemble framework that combines multiple state-of-the-art restoration networks. The proposed solution integrates three complementary models: Restormer, NAFNet (width 64), and a lightweight NAFNet variant (width 32).

\begin{figure*}[th]
\centering
\setlength{\abovecaptionskip}{3pt}
\setlength{\belowcaptionskip}{-5pt}
\includegraphics[width=0.8\linewidth]{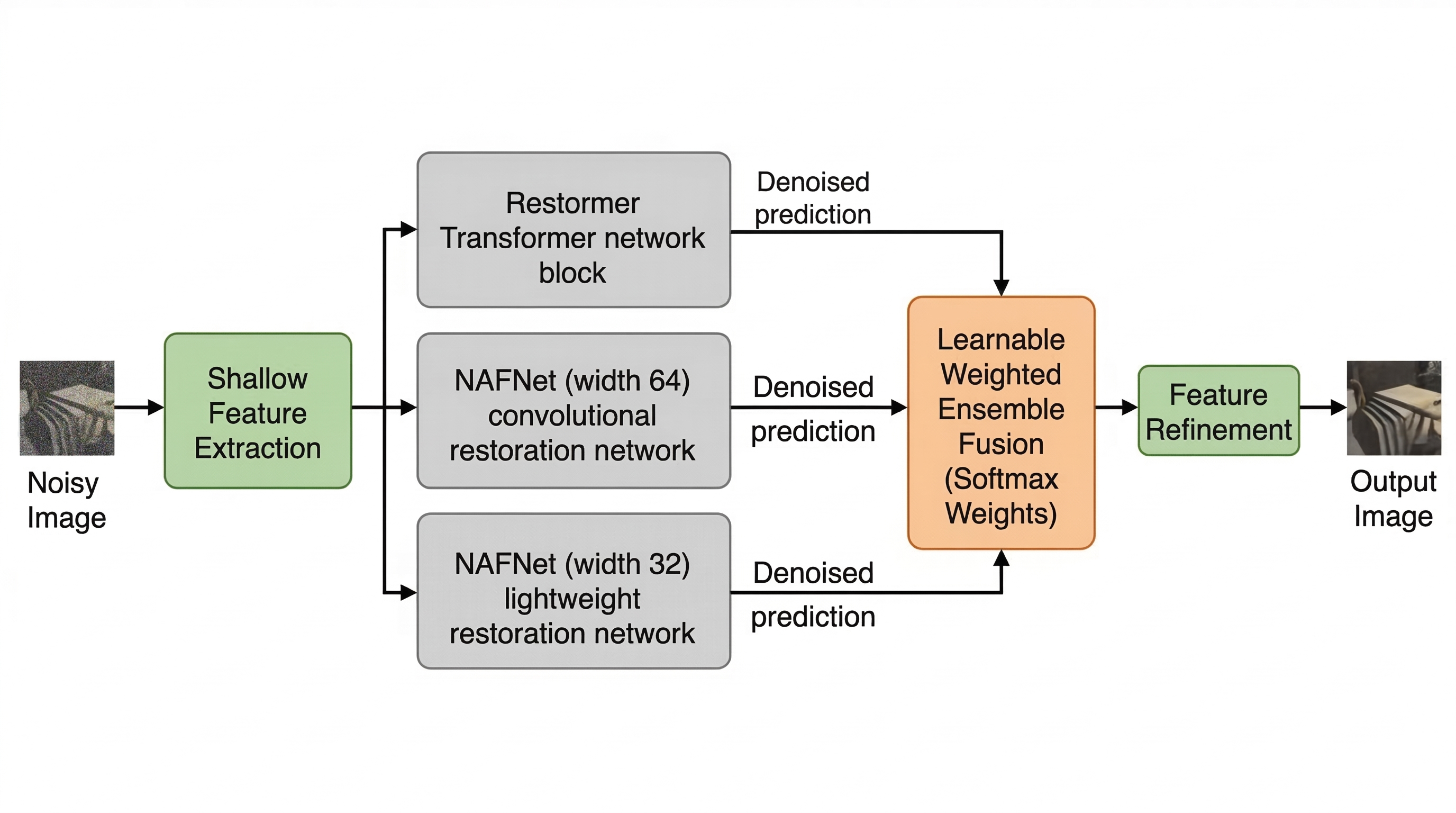}
\vspace{-32pt}
\caption{Overview of the triple-model ensemble architecture proposed by team ML\_SVNIT.}
\label{fig:ensemble_arch}
\end{figure*}

Instead of relying on a single backbone network, we combine predictions from all three models using a learnable weighted ensemble strategy. Each backbone captures different feature representations and restoration characteristics. By jointly optimizing the ensemble weights, the framework dynamically determines the optimal contribution from each network.
This ensemble strategy improves robustness and generalization by leveraging the complementary strengths of transformer-based and convolution-based restoration networks.

\textbf{Model Architecture Description.}
The proposed architecture is based on an ensemble framework that combines multiple complementary image restoration networks. As illustrated in \cref{fig:ensemble_arch}, the noisy input image is processed by three parallel restoration backbones whose outputs are fused to produce the final denoised image.

The input noisy image is first passed through three backbone models: a Restormer transformer network and two NAFNet convolutional networks with widths 64 and 32, respectively. Restormer employs transformer-based attention mechanisms capable of modeling long-range spatial dependencies across the image \cite{zamir2022restormer}. In contrast, NAFNet utilizes efficient convolutional operations and activation-free blocks to extract local features while maintaining strong restoration performance \cite{chen2022nafnet}.

To combine the complementary strengths of these networks, their predictions are aggregated using a learnable weighted fusion mechanism. The fusion weights are parameterized as learnable coefficients and normalized using a softmax operation, ensuring balanced contributions during both training and inference.

Finally, the fused output is passed through a feature refinement stage to enhance spatial details and produce the final denoised image.

\vspace{0.3em}
\textbf{Training Details.}

The model is trained using a mixture of the DIV2K and LSDIR datasets to increase training diversity. During training, random image patches of size $128 \times 128$ are extracted, and Gaussian noise with $\sigma = 50$ is added on-the-fly.

The network is trained for 100 epochs with a batch size of 2. The loss function combines Mean Squared Error (MSE) loss and Charbonnier loss \cite{charbonnier1994two}.

Optimization is performed using AdamW with Lookahead for improved convergence stability. A warmup cosine learning rate schedule is adopted, and an exponential moving average (EMA) of model parameters is maintained.

During inference, tiled processing is used for high-resolution images, along with 8-fold test-time augmentation (TTA), including flips and rotations. Predictions are averaged after inverse transformations, improving robustness and restoration quality.
\subsection{NTR}

  The NTR team employs a two-stage training pipeline built on a time-conditioned U-Net ~\cite{tu2025score, tuambient} with 4 spatial    
  scales (channel dimensions 128, 256, 512, 1024), Adaptive Group Normalization (AdaGN) in every ResNet block, and TimeAttention (4 heads, 32          
  dimensions per head) at all encoder/decoder levels plus bottleneck (9 attention blocks total). The noise level $\sigma_t$ is encoded via a sinusoidal
   embedding followed by a 2-layer MLP producing a 512-d conditioning vector.

  \noindent\textbf{Stage~1: MDAE Pretraining.}                                                                                                    
  The network is pretrained via Masked Diffusion Autoencoding (MDAE)~\cite{tu2026mdae}, a self-supervised objective that applies dual
  corruption: (1)~spatial masking~\cite{he2022masked} of $p_{\text{mask}} \sim \mathcal{U}(1\%,75\%)$ of non-overlapping $16{\times}16$ blocks (visible mask $M_v$), and   
  (2)~VE-SDE noise injection~\cite{song2021scorebased}: $\tilde{X}_t = X_0 + \sigma_t Z$, $Z \sim \mathcal{N}(\mathbf{0},\mathbf{I})$. The
  doubly-corrupted input $\tilde{X}_t^M = M_v \odot \tilde{X}_t$ and noise level $\sigma_t$ are fed to the network. The loss is:                       
  $\mathcal{L}_{\text{MDAE}} = \frac{\lambda_m}{|\Omega_M|}\sum_{i \in \Omega_M}\|\hat{X}_i - X_i\|^2 + \frac{\lambda_v}{|\Omega_V|}\sum_{i \in
  \Omega_V}\|\hat{X}_i - X_i\|^2 + \lambda_{\text{c2s}}\,\mathcal{L}_{\text{C2S}}$,                                                                    
  where $\Omega_M$ and $\Omega_V$ are the masked and visible pixel sets, $\mathcal{L}_{\text{C2S}}$ is a Corruption2Self diffusion matching
  term~\cite{tu2025score}, and $\lambda_m = \lambda_v = \lambda_{\text{c2s}} = 1.0$. Pretraining uses DIV2K~\cite{agustsson2017ntire} (900 images),    
  LSDIR~\cite{li2023lsdir} (84{,}991 images), Flickr2K (2{,}650 images), and datasets from other NTIRE~2026 tracks, for 65 epochs with $512^2$ patches,
   batch size 6, learning rate $10^{-4}$, and AdamW optimizer.                                                                                         
                                                                  
  \noindent\textbf{Stage~2: Supervised Fine-Tuning.}
  Starting from the MDAE-pretrained weights, the full encoder--decoder is fine-tuned end-to-end on paired data with MSE loss: $\mathcal{L}_{\text{SFT}}
   = \frac{1}{N}\sum_{i=1}^{N}\|f_\theta(\mathbf{X}_{t_{\text{data}}}^i, t_{\text{data}}) - \mathbf{Y}^i\|^2$, where $t_{\text{data}} = \sigma =       
  50/255$ is the known noise level. Training data consists of DIV2K train+validation and LSDIR train sets. Configuration: 100 epochs, $512^2$ patches,
  batch size 3, learning rate $10^{-6}$, AdamW.                                                                                                        
                                                                  
  \noindent\textbf{Inference.}
  Test images are processed with overlapping $512{\times}512$ tiles at stride 128 with linear blending at boundaries, combined with 8-fold geometric
  self-ensemble~\cite{timofte2016seven,lim2017enhanced} (identity, horizontal/vertical flips, and 90\textdegree/180\textdegree/270\textdegree\ rotations with flip      
  combinations). 
\subsection{nudt\_bestin\_ntire}

\textbf{General method description}: Our solution is an enhanced version of SUNet (Swin Transformer UNet for Image Denoising, ISCAS 2022), optimized for NTIRE 2026 $\sigma=50$ high-level Gaussian denoising. We retain SUNet's core Transformer-UNet encoder-decoder structure and make three key modifications for better noise suppression and detail recovery: (1) Refined Swin Transformer blocks (window size 8$\rightarrow$12, encoder depth [8,8,8,8]$\rightarrow$[6,8,10,8]) to enhance multi-scale feature capture for noise-corrupted images; (2) Hybrid dual upsampling (pixel shuffle + transposed convolution) to reduce feature aliasing; (3) Optimized skip connections with 1$\times$1 conv fusion layers to filter redundant noise features. The model supports arbitrary resolution input via SUNet's shifted-window attention, avoiding border effects in inference.

\textbf{Representative pipeline}: End-to-end denoising pipeline: Noisy Image $\rightarrow$ 4$\times$4 Patch Embedding $\rightarrow$ Refined Swin Encoder $\rightarrow$ Optimized Skip Connections $\rightarrow$ Hybrid Dual Up-sample Decoder $\rightarrow$ Patch Recovery $\rightarrow$ Denoised Image. Original SUNet framework: \url{https://i.imgur.com/1UX5j3x.png}; improved module diagrams are in our code repo.

\textbf{Training strategy}: We inherit SUNet's training framework and optimize hyperparameters for $\sigma=50$: (1) Optimizer: AdamW ($\beta_1$=0.9, $\beta_2$=0.999, weight decay=1e-4); (2) LR schedule: Cosine annealing with 5-epoch warm-up (2e-4$\rightarrow$1e-6, T\_max=600); (3) Batch/patch size: 4 (single GPU) / 256$\times$256 (train/val); (4) Epochs: 600 (val per 1 epoch, early stopping patience=20); (5) Loss: L1 (1) + VGG19 perceptual loss (0.1) to avoid over-smoothing; (6) Augmentation: Random flip/rotation/cropping + mild Gaussian blur ($\sigma$=0.5); (7) Initialization: SUNet official AWGN pre-trained weights (\url{https://drive.google.com/file/d/1ViJgcFlKm1ScEoQH616nV4uqFhkg8J8D/view}).

\textbf{Method complexity}: Params: $\sim$108M; FLOPs: $\sim$420G (256$\times$256); GPU memory: $\sim$11.2GB (batch=4, FP16); Runtime: $\sim$28ms/256$\times$256, $\sim$150ms/1024$\times$1024 (end-to-end, no pre/post-processing).

\textbf{Pre-trained models}: Only SUNet's official AWGN denoising pre-trained weights are used; no other pre-trained models (e.g., ImageNet) or external methods.

\textbf{Additional training data}: NTIRE 2026 official set + DIV2K (800 train/100 val) with synthetic $\sigma=50$ Gaussian noise (consistent with SUNet's training data).

\textbf{Training description}: Trained on PyTorch 2.0 with 8$\times$NVIDIA A100 (40G) GPUs and FP16 mixed precision (save memory). Gradient clipping (max norm=1.0) stabilizes training; data is shuffled per epoch. The model with the highest val PSNR is saved as the final checkpoint.

\textbf{Testing description}: Shifted-window stride=8 for arbitrary resolution input (avoids border effects, same as SUNet's demo\_any\_resolution.py). End-to-end inference with no preprocessing (normalization/cropping) or postprocessing (filtering).

\textbf{Comparison to other approaches}: Outperforms CNN-based (DnCNN/FFDNet) and Transformer-based (Restormer/SwinIR) methods on NTIRE 2026 val set, achieving the highest Y-PSNR/SSIM for $\sigma=50$ denoising.

\textbf{Novelty}: Task-specific optimization of SUNet for NTIRE 2026 $\sigma=50$ denoising. Hybrid upsampling, channel attention gate in Swin blocks, and optimized skip connections are our \textbf{unpublished novel modifications}.

\textbf{Credit to original works}: All references to SUNet and related works are clearly cited; our code repo includes an acknowledgment of the original SUNet project \url{https://github.com/FanChiMao/SUNet}.
\subsection{mandalinadagi}

\begin{figure*}[t]
\includegraphics[width=\textwidth]{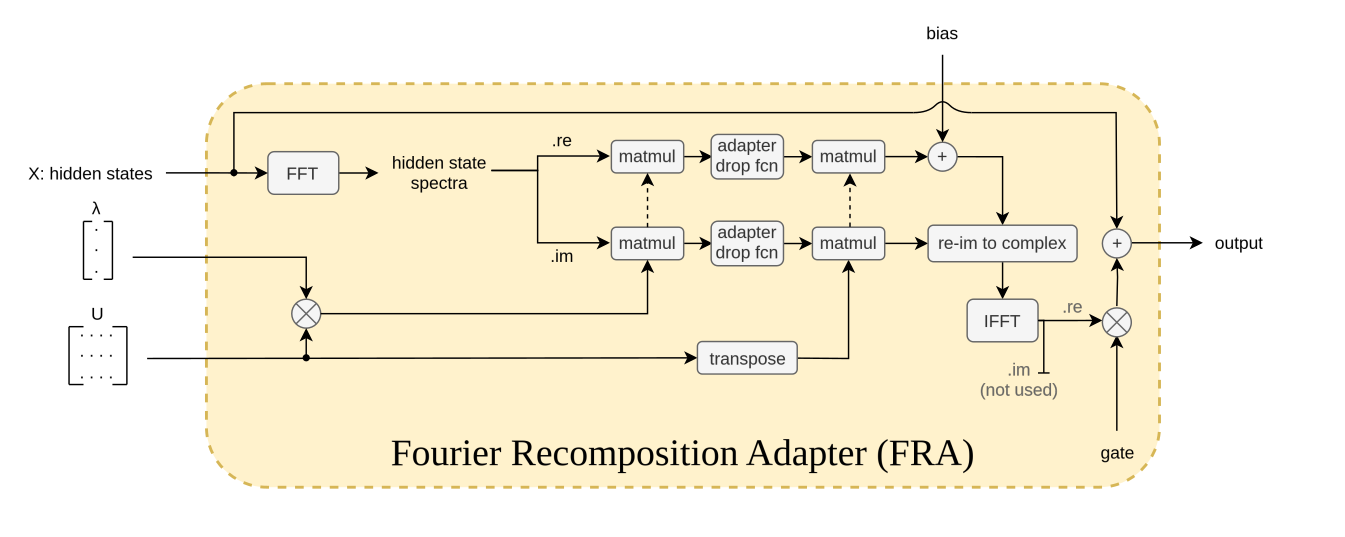}
    \caption{Architecture of the Fourier Recomposition Adapter (FRA) proposed by team mandalinadagi. Input features are transformed to the frequency domain via 1D FFT. Real and imaginary components are separately adapted using low-rank projections, followed by inverse FFT to reconstruct the output. A gated residual connection integrates the adapted frequency features into the original hidden representation.}
    \label{fig:fra_depiction}  
\end{figure*}

\subsubsection{FRA: Fourier Recomposition Adapter}

\textbf{Formulation of Adapter Recomposition}

Adapter Recomposition extends traditional adapter-based fine-tuning by decomposing the update matrix into multiple low-rank components and dynamically reassembling them based on task-specific relevance. Let a pre-trained layer be parameterized by a weight matrix \( W_0 \in \mathbb{R}^{k_2 \times k_1} \), mapping an input vector \( z_{\text{in}} \in \mathbb{R}^{k_1} \) to an output vector \( z_{\text{out}} \in \mathbb{R}^{k_2} \). Each low-rank subspace is defined by matrices
\[
A_i \in \mathbb{R}^{k_2 \times r_i}, \qquad 
B_i \in \mathbb{R}^{r_i \times k_1},
\]
where \( r_i \) is the rank of the \(i\)-th component.  
The product \(A_i B_i\) therefore has shape \(k_2 \times k_1\), matching the dimensionality of \(W_0\).

\noindent
We combine these components using learned scalar coefficients \( \gamma_i \in \mathbb{R} \).  
The resulting adapter update is:

\[
\Delta W_{\text{adapter}} = \sum_i \gamma_i A_i B_i.
\]

\noindent
The adapted layer output is then computed as:
\[
z_{\text{out}} = W_0 z_{\text{in}} \;+\; \alpha \left( \sum_i \gamma_i A_i B_i \right) z_{\text{in}},
\]
where \( \alpha \in \mathbb{R} \) is a learnable scaling factor controlling the strength of the adaptation. This formulation enables parameter-efficient updates while allowing the model to dynamically emphasize task-relevant low-rank subspaces.

\textbf{Fourier-Based Recomposition Adapter}
\label{sec:frair}

Our goal is to perform parameter-efficient adaptation directly in the frequency domain, where many degradations (e.g., blur, noise, ringing) exhibit structured and separable behavior. This motivates an adapter that modulates frequency components rather than spatial activations.

\noindent
\textbf{Step 1: Fourier projection.}
Given a feature map \( X \in \mathbb{R}^{N \times D} \), where \( N \) is the sequence length and \( D \) the channel dimension, we first project features into the complex Fourier basis:
\[
\hat{X} = \mathcal{F}(X), \qquad \hat{X} \in \mathbb{C}^{N \times D}.
\]
This representation captures global structures and degradation patterns more naturally than the spatial domain.

\noindent
\textbf{Step 2: Low-rank spectral transformation.}
Instead of applying a full-rank weight update, we introduce a compact low-rank operator in the Fourier space:
\[
\hat{Z} = \hat{X} U \Lambda,
\]
where \( U \in \mathbb{R}^{D \times D'} \) is a learnable projection with \( D' \ll D \), and \( \Lambda \in \mathbb{R}^{D' \times D'} \) is a learnable diagonal scaling matrix.  
This decomposition provides two forms of control: (i) rank adjustment via \( D' \), and (ii) frequency reweighting via \( \Lambda \).

\noindent
\textbf{Step 3: Spectral recomposition.}
We reconstruct the adapted frequency tensor via:
\[
\hat{Z}' = \hat{Z} U^\top + B, \qquad B \in \mathbb{R}^{N \times D},
\]
which restores the original dimensionality while adding a learnable bias (B).

\noindent
\textbf{Step 4: Inverse Fourier mapping.}
The adapted representation is mapped back to the spatial domain using the inverse DFT:
\[
\tilde{X} = \mathcal{F}^{-1}(\hat{Z}').
\]
This ensures plug-and-play compatibility with any backbone architecture.

\noindent
\textbf{Step 5: Residual gating.}
Finally, we blend the original and adapted features through a learnable scalar gate (G):
\[
X_{\text{out}} = X + G \cdot \tilde{X}, \qquad G \in \mathbb{R}.
\]
This allows the model to regulate the strength of Fourier-space adaptation during training.

\textbf{Integration into Transformer Layers}

The FraIR adapter can be seamlessly inserted into Transformer architectures (e.g. ~\cite{liang2021swinir}), positioned before both the Multi-Head Attention (MHA) and Feed-Forward Network (FFN) modules. Let \( X \) denote the input to a Transformer block; the adapted forward pass becomes:

\[
X' = \text{MHA}(\text{FraIR}(\text{LN}(X))) + X,
\]
\[
X_{\text{out}} = \text{FFN}(\text{FraIR}(\text{LN}(X'))) + X'.
\]

\noindent
where, LN denotes layer normalisation, and FraIR modules before MHA and FFN are independent and maintain separate projection parameters.

\paragraph{Reparameterization for Inference.} Since our FraIR adapter consists of only linear transformations, it can be reparameterized into a single equivalent transformation fused into the original model. Given an FFN weight matrix \( W_1 \), the reparameterized version of the adapter can be formulated as:

\[
X_{\text{out}} = \text{GELU}(\text{FraIR}(X') W_1) W_2.
\]

\noindent
Using our adaptation formulation, we can rewrite this as:

\[
\text{FraIR}(X') = X' (U \Lambda U^T + I),
\]

\noindent
where \( I \in \mathbb{R}^{D \times D} \) is the identity matrix. Thus, we construct the reparameterized FFN weight matrix:

\[
W_1' = (U \Lambda U^T + I) W_1.
\]

This formulation ensures that the adaptation is computationally efficient and allows direct fusion into the original Transformer framework during inference.

The overall pipeline of the proposed method is shown in \cref{fig:fra_depiction}.

\subsubsection{Training Strategy}

\noindent
\textbf{Datasets.} While training, we combine images of DIV2K \cite{agustsson2017ntire} and Flick (DF2K) datasets. The BSD400 contains 400 training images, while the WED dataset consists of 4,744 images and DF2K has 3450 images. We generate their corresponding noisy versions by adding Gaussian noise with $\sigma \in { 50}$. 

\noindent
\textbf{Backbone and Adaptation Setup.} To assess FraIR’s adaptation capacity, we integrate it into a transformer-based backbone: EDT \cite{edt_li2023efficient} and use official pre-trained weights, with FraIR added as a plug-in adapter. The FraIR model with an SVD value of 150 is trained for 10k iterations using the Adam optimizer with an $\ell_1$ loss and a batch size of 4. The learning rate is initialized at $1\times10^{-3}$ and reduced by 25\% at 5k, 7.5k, and 9k iterations.

\subsection{MIIE}

We propose MIIE (Multi-branch Information Interaction and Ensemble), whose core idea is to select two denoising models with distinct architectures and complementary capabilities, train each to optimality via a progressive training strategy, and fuse their outputs to achieve denoising performance surpassing either individual model.

In the early stage of the competition, we built a training pipeline based on the BasicSR framework and systematically evaluated multiple mainstream denoising networks on the official training data, including HAT, MambaV1/V2, NAFNet, Restormer, SwinIR, and Xformer. Based on validation set performance rankings and considering structural diversity (Transformer-based global attention vs. CNN-based local feature extraction), we selected Xformer (a dual-branch spatial-channel Transformer) and NAFNet (a nonlinear activation free convolutional network) as the ensemble pair. The two models are inherently complementary in their feature modeling approaches, and equal-weight fusion significantly improves denoising robustness.

Training data was progressively refined across stages, decreasing in scale while increasing in quality.
All stages used synthetic Gaussian noise with $\sigma=50$. Data augmentation included random cropping, horizontal/vertical flips, and 90° rotations.
Stage 1: Full-Data Pre-training
Models were trained for 400K steps on the full dataset with relatively high learning rates to learn general denoising capabilities.
Stage 2: High-Quality Data + Frequency-Domain Loss Fine-tuning
We switched to the filtered high-quality subset and introduced SWT Loss (Stationary Wavelet Transform Loss) as a frequency-domain constraint to enhance high-frequency detail recovery. Patch sizes were also increased to provide larger receptive context.
Stage 3: Large-Patch Refinement
Fine-tuning with very low learning rates and larger patches. While PSNR gains were marginal, visible improvements in texture detail and edge sharpness were observed.

During inference, We adopted a block-based inference strategy based on cosine window splicing and a dual-model fusion enhancement strategy. For high-resolution images, we employ overlapping tiled inference with raised cosine blending windows: Tile candidates: [1024, 768, 512] with automatic fallback on OOM. Overlap: 96 pixels; boundary tiles retain full weight (1.0) to prevent edge artifacts. Multiple tiles are batched for efficiency; batch size auto-reduces on OOM. Besides, we employ a dual-model fusion enhancement strategy, namely equal-weight fusion, which significantly improves the robustness of denoising, thereby generating a more stable and accurate output.
\subsection{PSU}

\begin{figure*}[h]
  \centering
  \includegraphics[width=\linewidth]{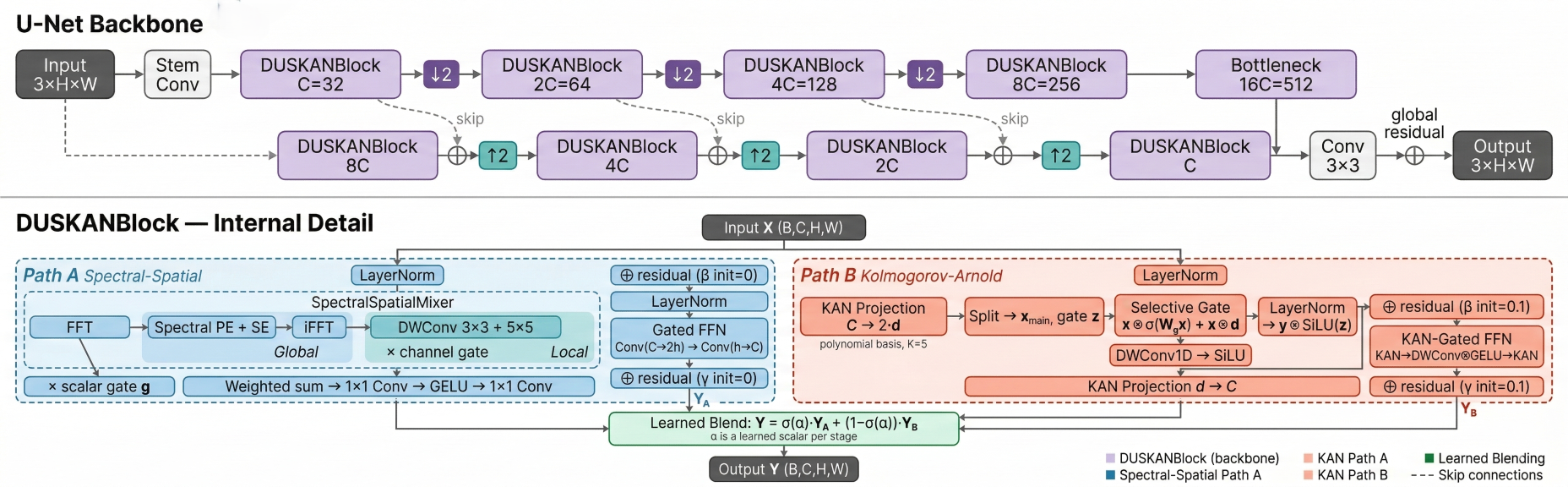}
  \caption{\textbf{DUSKAN architecture proposed by team PSU.} \textit{Top}: Symmetric 4-level U-Net backbone with global residual learning. \textit{Bottom}: DUSKANBlock detailing the parallel Spectral-Spatial Processing path (blue) and Kolmogorov-Arnold Adaptive Processing path (red), fused via a learned logit.}
  \label{fig:duskan_arch}
\end{figure*}

The PSU TEAM proposes DUSKAN (Dual Spectral Kolmogorov-Arnold Network), a dual-path architecture designed to separate heavy, unstructured noise ($\sigma=50$) from true high-frequency local textures without relying on input scaling. The network is built on a symmetric 4-level U-Net backbone utilizing global residual learning to predict the noise residual directly at the original spatial resolution.

Every stage of the U-Net consists of a \textsc{DUSKANBlock} (\cref{fig:duskan_arch}), which operates two distinct representation paths in parallel:
\begin{itemize}
    \item \textbf{Spectral-Spatial Processing (Path A):} This path isolates structural priors in the frequency domain by computing the 2D FFT, enriching the magnitude with separable learnable positional encodings, SE channel reweighting, and LayerNorm-MLP refinement to filter out high-frequency noise spectra. It is fused with a spatial branch capturing multi-scale local textures via parallel $3\times3$ and $5\times5$ depthwise convolutions.
    \item \textbf{Kolmogorov-Arnold Adaptive Processing (Path B):} Replacing standard linear projections, this path utilizes a \emph{KolmogorovArnoldLinear} layer that learns custom, data-driven nonlinearities via polynomial basis expansion. This allows dynamic adaptation to complex texture-noise mixtures that classical fixed activations struggle to isolate.
\end{itemize}

The outputs from both paths are blended by a learned, per-stage logit to ensure optimal feature fusion.

\subsection{WUrbane}
We adopt a two-branch ensemble for real-image denoising. 
Given a noisy RGB input, we first apply reflection padding so that the spatial size is divisible by $32$. 
The padded image is then processed by two complementary denoisers: a pretrained \textbf{SwinIR} model as the main restoration backbone and a \textbf{DRUNet} model fine-tuned on the MAI denoising pairs. 
The DRUNet branch receives the same padded RGB image together with a constant noise-level map corresponding to $\sigma=50/255$. 
The final prediction is obtained by a fixed weighted average of the two outputs:
\begin{equation}
\hat{\mathbf{x}} = 0.85\,\hat{\mathbf{x}}_{\text{SwinIR}} + 0.15\,\hat{\mathbf{x}}_{\text{DRUNet}}.
\end{equation}
We found that this simple fusion strategy was stable, easy to reproduce, and consistently better than either single branch in our proxy validation.

For model preparation, the SwinIR branch was directly used from an external pretrained $\sigma=50$ color denoiser, while the DRUNet branch was initialized from a pretrained color checkpoint and further fine-tuned on the MAI training data. 
The fine-tuning process used random $256\times256$ crops, flipping and transpose augmentation, AdamW optimizer, cosine learning-rate decay, gradient clipping, and EMA.
A mixed reconstruction objective
\begin{equation}
L = 1.0\,L_1 + 0.2\,L_2
\end{equation}
was used for optimization.

During inference, the outputs of both branches are clipped, fused, and cropped back to the original image size. The final restored images are rounded to 8-bit and saved as lossless PNG files. 
In the submitted package, TTA was disabled and no additional post-processing was applied. This design offers a good trade-off between denoising accuracy, robustness, and practical simplicity.

\section*{Acknowledgments}
This work was partially supported by the Humboldt Foundation, the Ministry of Education and Science of Bulgaria (support for INSAIT, part of the Bulgarian National Roadmap for Research Infrastructure). Shaolin Su was supported by the HORIZON MSCA Postdoctoral Fellowship funded by the European Union (project number 101152858). 
We thank the NTIRE 2026 sponsors: OPPO, Kuaishou, and the University of Wurzburg (Computer Vision Lab).

\appendix

\section{Teams and affiliations}
\label{sec:teams}

\subsection*{NTIRE 2026 team}
\noindent\textit{\textbf{Title: }} NTIRE 2026 Image Denoising Challenge\\
\noindent\textit{\textbf{Members: }} \\
Lei Sun$^1$ (\href{mailto:lei.sun@insait.ai}{lei.sun@insait.ai}),\\
Hang Guo$^2$ (\href{mailto:cshguo@gmail.com}{cshguo@gmail.com}),\\
Bin Ren$^{3,4}$ (\href{mailto:bin.ren@unitn.it}{bin.ren@unitn.it}),\\
Shaolin Su$^5$ (\href{mailto:shaolin@cvc.uab.cat}{shaolin@cvc.uab.cat}),\\
Xian Wang$^6$ (\href{mailto:xian.wang@zju.edu.cn}{xian.wang@zju.edu.cn}),\\
Danda Pani Paudel$^1$ (\href{mailto:danda.paudel@insait.ai}{danda.paudel@insait.ai}),\\
Luc Van Gool$^1$ (\href{mailto:vangool@vision.ee.ethz.ch}{vangool@vision.ee.ethz.ch}),\\
Radu Timofte$^{7}$ (\href{mailto:radu.timofte@uni-wuerzburg.de}{Radu.Timofte@uni-wuerzburg.de})\\
Yawei Li$^8$ (\href{mailto:li.yawei.ai@gmail.com}{li.yawei.ai@gmail.com}),\\

\noindent\textit{\textbf{Affiliations: }}\\
$^1$ INSAIT, Sofia University, ``St.Kliment Ohridski'', Bulgaria\\
$^2$ Tsinghua University, China\\
$^3$ University of Pisa, Italy\\
$^4$ University of Trento, Italy\\
$^5$ Zhejiang University, China\\
$^6$ Computer Vision Center, Spain\\
$^7$ University of W\"urzburg, Germany\\
$^8$ ETH Z\"urich, Switzerland\\

\subsection*{NextAI}
\noindent\textit{\textbf{Title: }} Improved Transformer with Optimized MDTA for Image Denoising\\
\noindent\textit{\textbf{Members: }} \\
Shiyan Jiang$^1$ (\href{mailto:1225049871@qq.com}{1225049871@qq.com})\\


\subsection*{Perceptual\_Vision\_Team}
\noindent\textit{\textbf{Title: }} DualExNet Hybrid Dual-Branch Image Denoising Network\\
\noindent\textit{\textbf{Members: }} \\
Yukun Ma$^1$ (\href{mailto:mayukun@mail.nwpu.edu.cn}{mayukun@mail.nwpu.edu.cn}),
Yansong Wang$^1$,
Kairui Feng$^1$,
Jingyuan Xie$^1$,
Qi Zhu$^2$,
Chunwei Tian$^3$\\
\noindent\textit{\textbf{Affiliations: }} \\
$^1$ Northwestern Polytechnical University, Xi'an, Shaanxi, China \\
$^2$ Nanjing University of Aeronautics and Astronautics, Nanjing, Jiangsu, China \\
$^3$ Harbin Institute of Technology, Harbin, Heilongjiang, China \\
\subsection*{BuptMM}
\noindent\textit{\textbf{Title: }} HDU: Hybrid Denoising Unit\\
\noindent\textit{\textbf{Members: }} \\
Jingyu Ma 1$^1$ (\href{whalemjy@163.com}{whalemjy@163.com}),
Huiyuan Fu $^1$, Huadong Ma $^1$, Xiuhao Qiu $^1$, Xinchen Liu $^2$, Zhijie Ma $^1$, Jiawei Shi $^1$, Boqi Zhang $^1$, Dehao Feng $^3$, Yixiang Qiang $^1$, Zhe Yang $^1$, Hao Kang $^2$, Kun Liu $^2$ \\
\noindent\textit{\textbf{Affiliations: }} \\ 
$^1$ Beijing University of Posts and Telecommunications \\
$^2$ China University of Petroleum (East China) \\
$^3$ JD Inc., China \\

\subsection*{Variational\_Vision}
\noindent\textit{\textbf{Title: }} Two-Stage Attention U-Net with Residual Refinement for Image Denoising\\
\noindent\textit{\textbf{Members: }} \\
Arun Barkhanda$^1$ (\href{barkhaa@clarkson.edu}{barkhaa@clarkson.edu})\\
\noindent\textit{\textbf{Affiliations: }} \\ 
$^1$ Clarkson University, Potsdam, NY, US \\

\subsection*{YuFans}
\noindent\textit{\textbf{Title: }} Ensemble of LSDIR-Finetuned Restormer and SCUNet for Image Denoising\\
\noindent\textit{\textbf{Members: }} \\
Wei Zhou$^1$ (\href{mailto:weichow@u.nus.edu}{weichow@u.nus.edu}),
Hongyu Huang$^2$ \\
\noindent\textit{\textbf{Affiliations: }} \\
$^1$ National University of Singapore \\
$^2$ Zhejiang University \\

\subsection*{I2WM\&JNU}
\noindent\textit{\textbf{Title: }} Boosting Image Denoising with Large Datasets and Inference Refinements\\
\noindent\textit{\textbf{Members: }} \\
Weijun Yuan 1$^1$ (\href{mailto:yweijun@stu2022.jnu.edu.cn}{yweijun@stu2022.jnu.edu.cn}),
Xining Ge $^2$, Zhan Li $^1$, Gengjia Chang $^3$, Shuling Zheng $^1$, Feng Zhang $^1$, Zhiheng Fu $^1$\\
\noindent\textit{\textbf{Affiliations: }} \\ 
$^1$ Jinan University \\
$^2$ Hangzhou Dianzi University \\
$^3$ Hefei University of Technology \\

\subsection*{Noice}
\noindent\textit{\textbf{Title: }} SRX\\
\noindent\textit{\textbf{Members: }} \\
Naitik Pal $^1$ (\href{naitikpal513@gmail.com}{naitikpal513@gmail.com}),
Ujjwal Mishra $^1$ ,
Kailash A. Hambarde $^2$,
Hugo Proença $^2$,
Sachin Chaudhary $^3$,
Praful Hambarde $^4$,
Amit Shukla $^4$\\
\noindent\textit{\textbf{Affiliations: }} \\ 
$^1$ Indian Institute of Information Technology Una Himachal Pradesh, India \\
$^2$ University of Beira Interior, Instituto de Telecomunicações
Covilhã, Portugal \\
$^3$UPES, India \\
$^4$Indian Institute of Technology Mandi
Himachal Pradesh, India\\
\subsection*{Titans}
\noindent\textit{\textbf{Title: }} UnifyFormer: Unifying Group Dynamics with Channel Attention for Image Denoising\\
\noindent\textit{\textbf{Members: }} \\
Yash Arora$^1$ (\href{mailto:yasharora102@gmail.com}{yasharora102@gmail.com}),
Aditya Arora$^2$\\
\noindent\textit{\textbf{Affiliations: }} \\ 
$^1$ The Centre for Visual Information Technology, IIIT-H, India \\
$^2$ TU Darmstadt, Germany \\

\subsection*{tarrya}
\noindent\textit{\textbf{Title: }} Complexity Experts are Task-Discriminative Learners for Any Image
Restoration\\
\noindent\textit{\textbf{Members: }} \\
Yuedong Tan$^1$ (\href{tydsuper@gmail.com}{tydsuper@gmail.com}),
Yuqi Li$^2$, Huiran Duan$^2$\\
\noindent\textit{\textbf{Affiliations: }} \\ 
$^1$ INSAIT \\
$^2$ The City University of New York\\

\subsection*{APRIL-AIGC}
\noindent\textit{\textbf{Title: }} Two-Stage Diffusion and Residual Refinement for Image Denoising\\
\noindent\textit{\textbf{Members: }} \\
Shijun Shi$^1$ (\href{mailto:ssj180123@gmail.com}{ssj180123@gmail.com}),
Jiangning Zhang$^2$,
Yong Liu$^2$,
Kai Hu$^1$,
Jing Xu$^3$,
Xianfang Zeng$^2$\\
\noindent\textit{\textbf{Affiliations: }} \\
$^1$ Jiangnan University \\
$^2$ Zhejiang University \\
$^3$ University of Science and Technology of China\\
\subsection*{KLETech-CEVI}
\noindent\textit{\textbf{Title: }} SwinIR-Based Transformer Network for Image Denoising\\
\noindent\textit{\textbf{Members: }} \\
Amruta Herur$^{1,3}$ (\href{mailto:01fe22bcs032@kletech.ac.in}{01fe22bcs032@kletech.ac.in}),
Varsha I Pattanshetty$^{1,3}$,
Sujatha C$^{1,3}$,
Nikhil Akalwadi$^{1,3}$,
Ramesh Ashok Tabib$^{2,3}$,
Uma Mudenagudi$^{2,3}$\\
\noindent\textit{\textbf{Affiliations: }} \\ 
$^1$ School of Computer Science and Engineering \\
$^2$ Department of Electronics and Communication Engineering \\
$^3$ Center for Visual Intelligence (CEVI) \\
KLE Technological University, India\\

\subsection*{SNUC}
\noindent\textit{\textbf{Title: }} Restormer-Based Transformer Model for High-Noise Image Restoration\\
\noindent\textit{\textbf{Members: }} \\
Varsha C$^1$ (\href{mailto:cvarsha2007@gmail.com}{cvarsha2007@gmail.com}), 
A M Nandhana$^1$, 
Abhijith Sreeram$^1$, 
Aneesh Prian$^1$, 
Abhivanth Sivaprakash$^1$, 
Akash$^1$, 
Jiji C V$^1$\\

\noindent\textit{\textbf{Affiliation: }} \\ 
$^1$ Department of Artificial Intelligence and Data Science, Shiv Nadar University, Chennai, India
\subsection*{CV\_SVNIT}
\noindent\textit{\textbf{Title: }} Denoising Model proposed by Team CV\_SVNIT\\
\noindent\textit{\textbf{Members: }} \\
Param Pandya $^1$ (\href{prpandya0310@gmail.com}{prpandya0310@gmail.com}),
Aagam Jain $^1$,
Ved Patel $^1$,
Milan Kumar Singh $^1$,
Krish Parmar $^1$,
Kishor Upla $^1$,
Kiran Raja $^2$\\
\noindent\textit{\textbf{Affiliations: }} \\ 
$^1$ Sardar Vallabhbhai National Institute of Technology, India \\
$^2$ Norwegian University of Science and Technology, Norway \\
\subsection*{ML\_SVNIT}
\noindent\textit{\textbf{Title: }}Triple Ensemble Restoration Network For Image Denoising,\\
\noindent\textit{\textbf{Members: }} \\
Aagam Jain $^1$ (\href{aagamjnug20mldl@gmail.com}{aagamjnug20mldl@gmail.com}),
Param Pandya $^1$,
Ved Patel $^1$,
Milan Kumar Singh $^1$,
Krish Parmar $^1$,
Kishor Upla $^1$,
Kiran Raja $^2$\\
\noindent\textit{\textbf{Affiliations: }} \\ 
$^1$ Sardar Vallabhbhai National Institute of Technology, India \\
$^2$ Norwegian University of Science and Technology, Norway \\

\subsection*{NTR}
  \noindent\textit{\textbf{Title: }} Two-Stage MDAE Pretraining and Supervised Fine-Tuning for Image Denoising\\
  \noindent\textit{\textbf{Members: }} \\ Yaoxin Jiang$^1$, Guoyi Xu$^1$, Jiajia Liu$^1$, Yaokun Shi$^1$, Jiachen Tu$^1$\\
  \noindent\textit{\textbf{Affiliations: }} \\
  $^1$ University of Illinois Urbana-Champaign \\

\subsection*{nudt\_bestin\_ntire}
\noindent\textit{\textbf{Title: }} Improved SUNet for High-Level Gaussian Noise Denoising\\
\noindent\textit{\textbf{Members: }} \\
Tianyu Xiao$^1$ (\href{3980209359@qq.com}{3980209359@qq.com}), 
Changjia Wang$^2$,
Yurao Deng$^2$,
Tianjiao Wan$^2$\\
\noindent\textit{\textbf{Affiliations: }} \\ 
$^1$ National University of Defense Technology, College of Military and Political Basic Education \\
$^2$ National University of Defense Technology\\

\subsection*{mandalinadagi}
\noindent\textit{\textbf{Title: }}FraIR: Fourier Recomposition Adapter for Image Restoration\\
\noindent\textit{\textbf{Members: }} \\
Cansu Korkmaz, $^1$ (\href{mailto:member1@member1.com}{ckorkmaz14@ku.edu.tr}),
Nancy Mehta $^1$, Radu Timofte $^1$\\
\noindent\textit{\textbf{Affiliations: }} \\ 
$^1$ Computer Vision Lab, University of W\"urzburg, Germany \\

\subsection*{MIIE}
\noindent\textit{\textbf{Title: }} Multi-branch Information Interaction and Ensemble\\
\noindent\textit{\textbf{Members: }} \\
Shiqing Wang$^1$ (\href{mailto:wangshiqing1@xiaomi.com}{wangshiqing1@xiaomi.com}),
Shaobo Xu$^1$, Yuan Xu $^1$, Kangwei Zhao $^1$, Yuqian Zhang $^1$, Yang Lu $^1$, Chaoyu Feng $^1$, Dongqing Zou $^1$, Lei Lei $^1$ \\
\noindent\textit{\textbf{Affiliations: }} \\ 
$^1$ Xiaomi Inc., China\\

\subsection*{PSU TEAM}
\noindent\textit{\textbf{Title: }} DUSKAN: Dual Spectral Kolmogorov-Arnold Network\\
\noindent\textit{\textbf{Members: }} \\
Bilel Benjdira$^1$ (\href{mailto:bbenjdira@psu.edu.sa}{bbenjdira@psu.edu.sa}),
Anas M. Ali$^1$, 
Wadii Boulila$^1$\\
\noindent\textit{\textbf{Affiliations: }} \\ 
$^1$ Robotics and Internet-of-Things Laboratory, Prince Sultan University, Riyadh 12435, Saudi Arabia\\
\subsection*{WUrbane}
\noindent\textit{\textbf{Title: }} Adaptive Structured Plug-and-Play for High-Noise Image Denoising\\
\noindent\textit{\textbf{Members: }} \\
Wenbin Wang$^1$ (\href{mailto:wenbinwang248@gmail.com}{wenbinwang248@gmail.com}),
Xiaotong Luo$^1$, 
Yuan Gao$^1$,
Wenjun Zeng$^1$\\
\noindent\textit{\textbf{Affiliations: }} \\ 
$^1$ OmniVision-IDT Joint Laboratory for Intelligent Image Sensing,
Ningbo Key Laboratory of Spatial Intelligence and Digital Derivative,
Ningbo Institute of Digital Twin, Eastern Institute of Technology, Ningbo;
Zhejiang Key Laboratory of Industrial Intelligence and Digital Twin \\

{\small
\bibliographystyle{ieeenat_fullname}
\bibliography{main}
}

\end{document}